# Efficient Planning under Uncertainty with Macro-actions

**Ruijie He** RUIJIE@CSAIL.MIT.EDU
*Computer Science and Artificial Intelligence Laboratory*
*Massachusetts Institute of Technology*
*Cambridge, MA 02139 USA*

**Emma Brunskill** EMMA@CS.BERKELEY.EDU
*Electrical Engineering and Computer Science Department*
*University of California, Berkeley*
*Berkeley, CA 94709 USA*

**Nicholas Roy** NICKROY@CSAIL.MIT.EDU
*Computer Science and Artificial Intelligence Laboratory*
*Massachusetts Institute of Technology*
*Cambridge, MA 02139 USA*

## Abstract

Deciding how to act in partially observable environments remains an active area of research. Identifying good sequences of decisions is particularly challenging when good control performance requires planning multiple steps into the future in domains with many states. Towards addressing this challenge, we present an online, forward-search algorithm called the Posterior Belief Distribution (PBD). PBD leverages a novel method for calculating the posterior distribution over beliefs that result after a sequence of actions is taken, given the set of observation sequences that could be received during this process. This method allows us to efficiently evaluate the expected reward of a sequence of primitive actions, which we refer to as macro-actions. We present a formal analysis of our approach, and examine its performance on two very large simulation experiments: scientific exploration and a target monitoring domain. We also demonstrate our algorithm being used to control a real robotic helicopter in a target monitoring experiment, which suggests that our approach has practical potential for planning in real-world, large partially observable domains where a multi-step lookahead is required to achieve good performance.

## 1. Introduction

Consider an autonomous helicopter tasked with protecting ships anchored in a busy harbor. At each time step, the helicopter must know if anything is moving too close to the ships it is guarding, but due to its sensor limits, the helicopter cannot observe the whole harbor at once. The only way to keep its ships safe is to keep moving continuously throughout the harbor, keeping track of all the other moving agents. The helicopter does well when it senses that another boat has moved too close to one of its charges, but false alarms are costly. The helicopter's controller must decide how to move around, what to report and when, in order to maximize its own performance.

This problem requires decision-making in an uncertain, partially observable domain, a common challenge for any agent operating in a real-world environment. The helicopter problem just described is an example of a general class of problems that are particularly difficult for two reasons. First, to make a decision, the agent must take into consideration its present estimate of the location and orientation of each of the targets. All of these quantities will typically be real-valued. In the standard terminology of Markov decision processes (MDPs), the state space consists of a large number of continuous variables. Second, to make a decision now, the agent must reason about how its estimate of the state of the world may change many time steps into the future, under different possible helicopter and target actions. Any problem with many variables to consider and a long time horizon to plan over suffers from the curse of dimensionality and the curse of history (Pineau, Gordon, & Thrun, 2003a). We refer to such problems as *large* and *long*.

In this paper we present a new planning algorithm for large, long, partially observable MDPs (POMDPs), such as the target monitoring example. Beyond target monitoring, there are numerous other problems, such as scientific exploration of extreme environments and autonomous management of retirement portfolios, which may be posed as large, long POMDPs.

Though there has been substantial progress in POMDP planning over the last decade, most approaches still struggle to scale to large domains described by many state variables, where each variable may take on a large or infinite number of potential values. Symbolic Perseus (Poupart, 2005) was used to find a good solution to a hand-washing domain with 11 state variables, but each variable took on a relatively small number of values (at most 10 values). Recently online forward search approaches have been used to achieve encouraging performance on some large[1] POMDPs, such as the work by Ross, Chaib-draa and Pineau (2008b) and Paquet, Tobin and Chaib-draa (2005). However, the cost of performing a generic forward search scales exponentially with the search horizon. The target monitoring example described above not only is too large to be solved by offline approaches, but, as we will demonstrate later, also requires a long horizon search to achieve good performance, limiting the effectiveness of standard forward search for long problems.

As an effort towards scaling to large, long, partially observable decision making, we introduce the Posterior Belief Distribution (PBD) algorithm. PBD leverages the insight that for certain environments which have specific structure, the distribution of belief states (which in turn are distributions over states) that arise from a fixed sequence of actions can be computed efficiently and analytically. This distribution over beliefs, or *posterior belief distribution*, allows us to scale to large, long POMDP problems using efficient forward search with temporally-extended action sequences, which we refer to as *macro-actions*. PBD selects an action for the current belief by planning over a restricted policy space defined by the input macro-action set, and then re-plans after the selected action is taken and a new observation is received. Note that this implies that the policy executed does not necessarily equal the policy space used for planning, since only the first step of a macro-action is executed before re-planning is performed. This characteristic of PBD is very similar to receding horizon controllers (RHC) (such as Mayne, Rawlings, Rao, & Scokaert, 2000; Kuwata & How, 2004). RHCs consider a finite-horizon policy space when performing planning, but can execute over a much longer horizon by repeatedly re-planning.

In this paper we demonstrate that our PBD algorithm achieves good performance on large, long POMDP problems which are either outside the scope of prior approaches, or on which prior approaches fail to find good quality policies. Our experimental results demonstrate that PBD performs well with an attractive computational cost on several large, long simulation problems, including a variant of the ROCKSAMPLE POMDP benchmark problem (Smith & Simmons, 2005) and a simulated target monitoring example. We also demonstrate the PBD algorithm on a real-world version of the target monitoring problem, where we use a robotic helicopter platform to monitor multiple ground vehicles (Section 6.4). This demonstration suggests that PBD has practical potential for real

1. Unless otherwise specified, when we describe a domain as “large” we will be referring to a domain described by the values of a number of state variables, where each variable can take on many or an infinite number of values.

robotic domains. In this paper, the macro-actions are assumed to be provided by a domain expert[2]; however, to decouple the impact of our specific choice of macro-actions, we also provide experimental results where we modify alternate approaches (including a state-of-the-art planner) to use macro-actions, and still find performance advantages for our presented methods.

The rest of the paper is organized as follows. Section 2 first provides a brief background on planning under uncertainty using forward search. We then introduce our PBD algorithm in Section 3, and consider a slight variant of PBD that is applicable to a larger set of domains in Section 4. In Section 5 we provide a formal analysis of the PBD algorithm, and then in Section 6 we present experimental results. We present related work in Section 7 and finally conclude in Section 8.

## 2. Background: Planning under Uncertainty using Forward Search

Formally, we assume that our decision-making under state-uncertainty problem consists of the following known components:

- $\mathcal{S}$ is a set of states. Each state $s \in \mathcal{S}$ consists of an assignment of values to each of $L$ state variables, $s_l$. The domain of each state variable may be either discrete or continuous.
- $\mathcal{A}$ is a set of actions (controls) $a \in \mathcal{A}$, which can be either discrete or continuous.
- $\mathcal{Z}$ is a set of observations $z \in \mathcal{Z}$, which can be either discrete or continuous.
- $p(s'|s, a)$ is a transition function (also known as a dynamics model) which encodes the probability of transitioning to state $s'$ after taking action $a$ from state $s$. We assume the dynamics satisfy the Markov assumption that the new state is only a function of the immediately prior state and action.
- $p(z|s)$ is an observation function (also known as a measurement or sensor model) that encodes the probability of receiving observation $z$ in state $s$.[3]
- $b_0$ is a distribution over possible initial states, where $b_0(s)$ is the probability that the initial state is $s$. This distribution is known as the initial belief state, and is a well-formed distribution that sums to one across all states.
- $r(s, a)$ is a reward (or cost) function that describes the utility the agent receives for taking action $a$ in state $s$. Slightly abusing notation, $r(b, a)$ is the expected reward for taking action $a$ given a distribution over current states (belief) $b$.
- $\gamma$ is a discount factor that determines the weights of immediate rewards relative to the rewards that will be received at a later time step.

The states $\mathcal{S}$ are not fully observable. Instead, at every time step, the agent receives an observation after taking an action. The agent must therefore make decisions based on the prior history of observations it has received, $z_{1:t}$, and actions it has taken, $a_{1:t}$, up to time $t$. As the world states are assumed to be Markov, instead of maintaining an ever-expanding list of past observations and

2. In other work we have demonstrated that we can automatically construct good macro-actions for smaller POMDPs (He, Brunskill, & Roy, 2010b). Integrating these two lines of work is an interesting area for future work but is outside the scope of this paper.
3. It is easy to extend our framework to allow the observation to depend on the prior state, action, and posterior state.

actions, a sufficient statistic, known as a belief $b_t(s)$, is used to summarize the probability of the world being in each state given its past history,

$$b_t(s) = Pr(s_t = s|a_0, z_1, \ldots, z_{t-1}, a_{t-1}, z_t). \tag{1}$$

The agent can therefore plan based only on the current belief state, rather than on all past actions and observations (Smallwood & Sondik, 1973). For example, in the target monitoring problem introduced in Section 1, the agent maintains a belief over the possible locations of each target. The agent updates its belief at each step, after taking an action $a$ and receiving an observation $z$ (such as a camera image of a far off target), using the Bayes filter:

$$b'(s') = \tau(b, a, z) = \eta \, p(z|a, s') \int_{s \in \mathcal{S}} p(s'|s, a) b(s) ds \tag{2}$$

where $\tau(b, a, z)$ represents the belief update function and $\eta$ is a normalization constant.

The planning problem is to compute a policy $\pi : b \to a$, which is a mapping from belief states to actions, that maximizes the expected sum of future[4] discounted utilities:

$$\pi = \text{argmax} \left[ \sum_{i=1}^{\infty} \gamma^i E[r(b_i)] \right], \tag{3}$$

where $E[r(b_i)]$ denotes the expected reward at time step $i$ given the actions specified by $\pi$ and possible observations received.

Many POMDP solvers, such as those by Smith and Simmons (2005), Porta, Vlassis, Spaan, and Poupart (2006) and Kurniawati, Hsu, and Lee (2008), perform POMDP planning offline by calculating a value function over the belief space $V : b \to \mathcal{R}$. $V(b)$ is the expected total reward of starting from any belief state $b$ and following an optimal policy[5],

$$V(b) = \max_{a \in \mathcal{A}} \big[ r(b, a) + \gamma \int_{z \in \mathcal{Z}} p(z|b, a) V(\tau(b, a, z)) \big], \tag{4}$$

where $p(z|b, a) = \int_s p(z|s, a) b(s) ds$. Given a value function over the belief space, a policy $\pi$ can be extracted by finding the action $a$ which maximizes Equation 4.

Instead of computing a value function over the entire belief space in advance of acting, we take an alternate approach of planning online, only explicitly computing a policy (that is, an action) for the current belief. In particular, an action is selected by performing a fixed-horizon forward search which is used to estimate the values of each of the possible action choices starting from the current belief. This action-selection approach is closely related to methods from the controls community, including Model Predictive/Receding Horizon Control, and forward search has also received recent attention in the AI POMDP community (see the recent survey in Ross, Pineau, Paquet, & Chaib-draa, 2008a).

To select an action for the current belief, generic forward search approaches compute a lookahead AND-OR tree (Figure 1). The goal of the tree is to estimate the value of taking each of the

4. We will assume in this paper that we are interested in problems with an infinite horizon. If the problem has a finite horizon, the discount factor $\gamma$ can be set to 1, and our forward search process (which we will shortly describe) will search out to a depth of at most the problem's finite horizon.
5. This is often intractable to compute, so in practice the value function is often approximate.

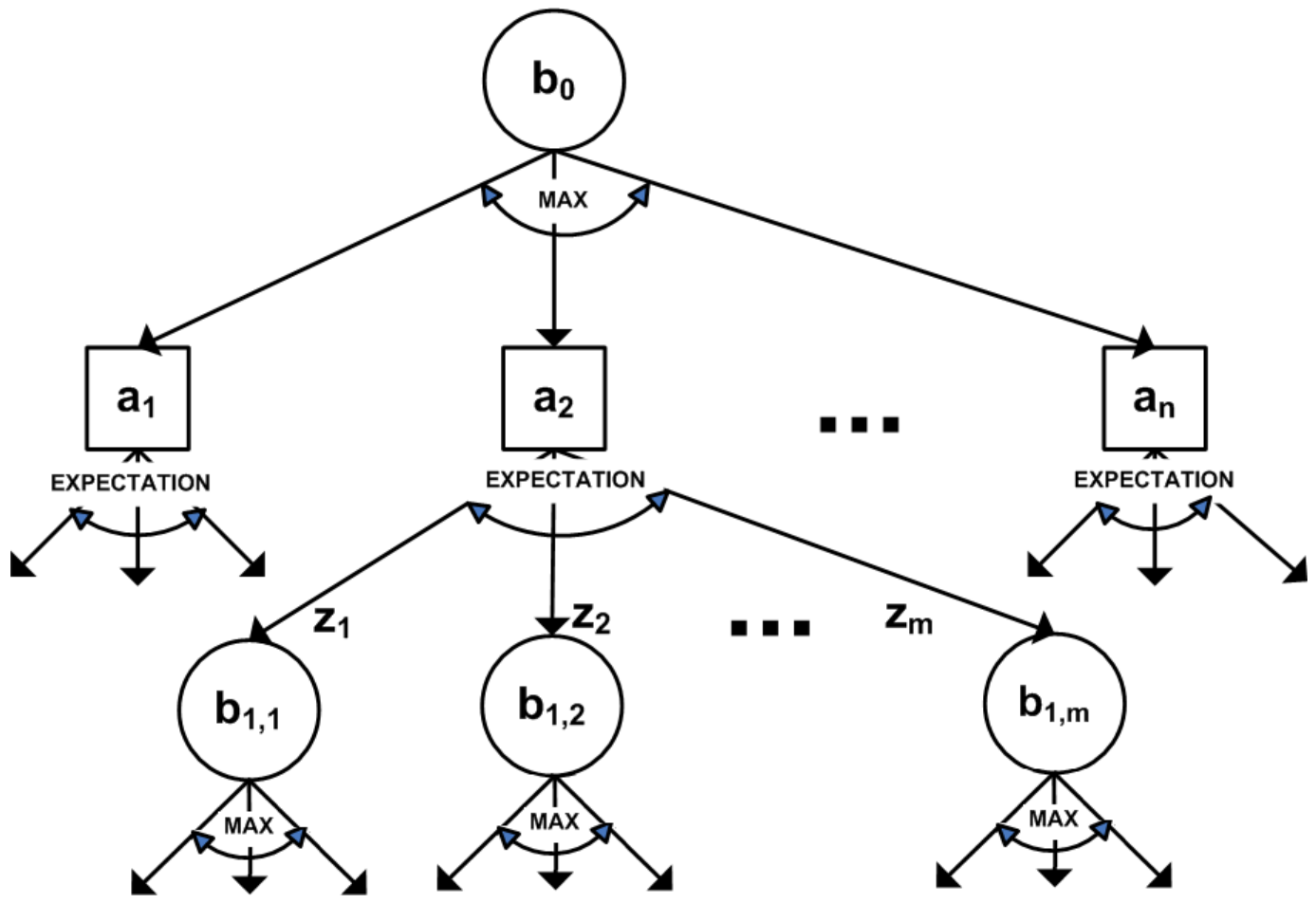

Figure 1: A forward search tree. $a$ are actions, $z$ are observations, and $b$ are beliefs. $b_0$ is the initial belief, while $b_{i,j}$ refers to the $j$th belief leaf node at depth $i$.

possible actions from the current belief $b$, in order to take the action with the greatest value. Given the root belief $b$, the tree is constructed by first branching on all possible actions from the root. After each action, the tree then branches on possible observations. For each distinct action-observation combination, we can compute the resulting internal belief that would occur if that action-observation trajectory were followed using Equation 2. This process of alternately branching on actions and observations is repeated out to a finite depth. This depth, known as the search horizon, determines how far into the future the effects of actions are considered when selecting a possible action for the root (current) belief state.

Once the tree has been constructed, the value of the actions at the root are computed by propagating the rewards from the beliefs at the leaf nodes back to the root. Starting at the leaf node rewards, we take an expectation over observations. We then add in the expected immediate reward from taking the parent action, and next take the maximum reward across all sibling action nodes. This process is repeated all the way up to the root node. The expected rewards are maximized across actions but summed across observations because the agent can choose which action to take, but must optimize over the expected distribution of observations.

After the planning phase, the forward search procedure executes the action at the root with the largest value, and then receives an observation. Given the previous belief, action taken, and observation received, a new belief is computed using Equation 2. The forward search planning process then repeats, with the new belief as the root node. Re-planning after every time step enables the agent to condition on the action selected and the actual observation received.

There are a number of attractive characteristics of an online, forward-search framework. First, computational effort is directed only towards belief states that are reachable from the current belief under different actions. This property enables a forward search planner to compute a meaningful policy in an arbitrarily large environment, since only a subset of the environment is relevant at any point. Second, online, forward-search fits well into systems that need good, time constrained solutions where a large amount of advance computation is not possible. Lastly, forward search does not have to compute an explicit representation of the value function, which can be an advantage in factored domains where belief updating and immediate expected reward calculations are relatively simple, but the value function itself is complex to represent.[6]

However, the computational cost of generic forward search will still scale with the cost of the belief updating and immediate expected reward calculations, multiplied by the number of tree nodes which grows exponentially with the search horizon. The costs of belief updating and calculating the immediate expected reward typically scale either linearly or exponentially with the number of state variables and the size of their respective domains, depending on the independence relations among the state variables. When the state variables are continuously-valued, and therefore take on an infinite number of values, we will typically need to employ some parametric or compressed representation in order to make these calculations tractable. The number of tree nodes scales exponentially with the horizon according to $\mathcal{O}((|\mathcal{A}||\mathcal{Z}|)^H)$, where $|\mathcal{A}|$ and $|\mathcal{Z}|$ are the number of actions and observations respectively and $H$ is the search horizon. Therefore, standard forward search approaches will typically struggle when there are many state variables and/or state variables with large domains and when a large $H$-step lookahead is necessary to achieve good performance.

One approach to accelerating planning over large, long horizon problems is to use temporally extended macro-actions, a technique that has been used successfully in fully observable settings for some years (Sutton, Precup, & Singh, 1999). There has been limited exploration of these ideas for partially observable settings (exceptions include those by Theocharous & Kaelbling, 2003; Hsiao, Lozano-Pérez, & Kaelbling, 2008; Kurniawati, Du, Hsu, & Lee, 2009). In our work we define a macro-action as a finite open-loop sequence of primitive actions that is executed without regard to the observations received during the execution of this action sequence. For example, in our target monitoring problem, one macro-action could be for the helicopter to travel to a key region, which might involve a sequence of individual turns and straight line moves. By restricting the action space to a set of length $L$ macro-actions, the number of expanded nodes due to the action branching factor can be reduced from$|\mathcal{A}|^H$ to $|\tilde{\mathcal{A}}|^{\tilde{H}}$ where $\tilde{\mathcal{A}}$ is the set of length $L$ (or longer) macro-actions, and $\tilde{H} = \frac{H}{L}$ is the macro-action horizon or depth[7].

## 2.1 Macro-action Construction

If only a small set of macro-actions are evaluated during the search, the restricted action space will result in significant computational savings due to the smaller exponent $\tilde{H}$ (vs. $H$) in the computational complexity expression. However, this restriction can also result in poor algorithmic performance if all the macro-actions being evaluated are unsuitable. In this paper, we assume that macro-actions are provided by a domain expert as part of a comprehensive strategy to scaling up to large problems with a multi-step lookahead. The macro-actions we use in our experimental results consist of open-loop policies which are a function of properties of the belief state at which the macro-action is originated, and can be either computed and stored offline or computed online at every timestep. Further details are provided in the experimental section.

Our reliance on domain knowledge in this paper is similar to prior work in the fully observable community that separately investigated the potential advantage of macro-actions before turning to

6. An example of such a domain is one in which the state space is a set of independent variables, but the reward is an aggregate function of these variables.
7. The macro-action depth refers to the number of macro-actions that are executed in sequence from the root belief node to the leaves.

the challenge of learning these macro-actions (see the work by Sutton et al., 1999 for an overview of one particular formalism). Although constructing macro-actions automatically is beyond the scope of this paper, we have presented in related work a domain-independent algorithm (PUMA) that automatically generates macro-actions for planning in partially observable domains (He et al., 2010b). Borrowing the notion of sub-goal states from the fully-observable planning literature (McGovern, 1998; Stolle & Precup, 2002), PUMA uses a heuristic that macro-actions can be designed to take the agent, under the fully-observable model, from a possible start state under the current belief to a sub-goal state. The PUMA algorithm was tested on variations of the experimental domains that are used in this paper, and we encourage the reader to refer to the above-mentioned paper for more details.

Regardless of how the set of macro-actions are generated, several key computational challenges remain to scale macro-action forward-search to large, long environments. First, recall the number of nodes in generic forward search scales as $\mathcal{O}(|\mathcal{A}|^H|\mathcal{Z}|^H)$. Using macro-actions reduces the first term in the product, but does not directly change the second term, so the number of tree nodes still is an exponential function of the search horizon $H$. Second, using macro-actions does not directly alleviate the cost of performing belief updates and expected reward computations at each tree node, and these computational costs can be substantial in large domains. The central contribution of our paper is a method for efficiently and analytically computing the result of a macro-action given any possible observation sequence received during its execution. This will allow us to use temporally-extended actions to scale to certain types of large, long POMDPs.

## 3. The Posterior Belief Distribution Algorithm

To plan with macro-actions in a forward search manner, we must compute the expected reward received during a macro-action, as well as the expected future value after taking that macro-action. The reward the planner can expect to receive from a macro-action is the expected sum of the rewards under each of the posterior beliefs the agent will reach after each action in the macro-action. However, the process is complicated by the fact that posterior the belief is also a result of receiving an observation. As the agent does not know which observations will be received during the macro-action, it cannot compute a single posterior belief reached during the macro-action, and therefore cannot compute the expected reward.

Of course, an easy solution is to consider all possible observations, and compute the expected reward of all possible beliefs that can result from all possible observations that could be received during a macro-action. By computing the expected reward at each observation node, the AND-OR tree constructed during forward search implicitly computes this expectation over all possible observation sequences. But, if computing the expected reward of a macro-action requires enumerating all possible observation sequences that could be experienced during execution, the evaluation of a macro-action will grow intractable quickly (see Figure 2(a)). The number of observation sequences to be considered will grow exponentially with the length of the macro-action, and enumerating all possible observations may not even be feasible in domains with continuous observations. One alternative may be to sample observation sequences for a given macro-action (Figure 2(b)), but sampling is likely to still be computationally intensive due to the per-sample cost of performing a belief update and expected reward calculation at each step of each sampled observation sequence.

We can avoid this computational burden by realizing that it is sometimes possible to analytically represent the distribution over posterior beliefs. For a given sequence of actions, what we need is the expected reward for those actions; if we cannot compute the distribution over states ahead of time, but *can* compute a distribution over state distributions, we can still compute the expected reward. A graphical depiction of this process is shown in Figure 2(c). By analytically computing a distribution over beliefs, we avoid not only the exponential explosion of potential observation sequences (as a function of the macro-action length), but also the costly step of performing many individual belief updates along the possible observation sequences.

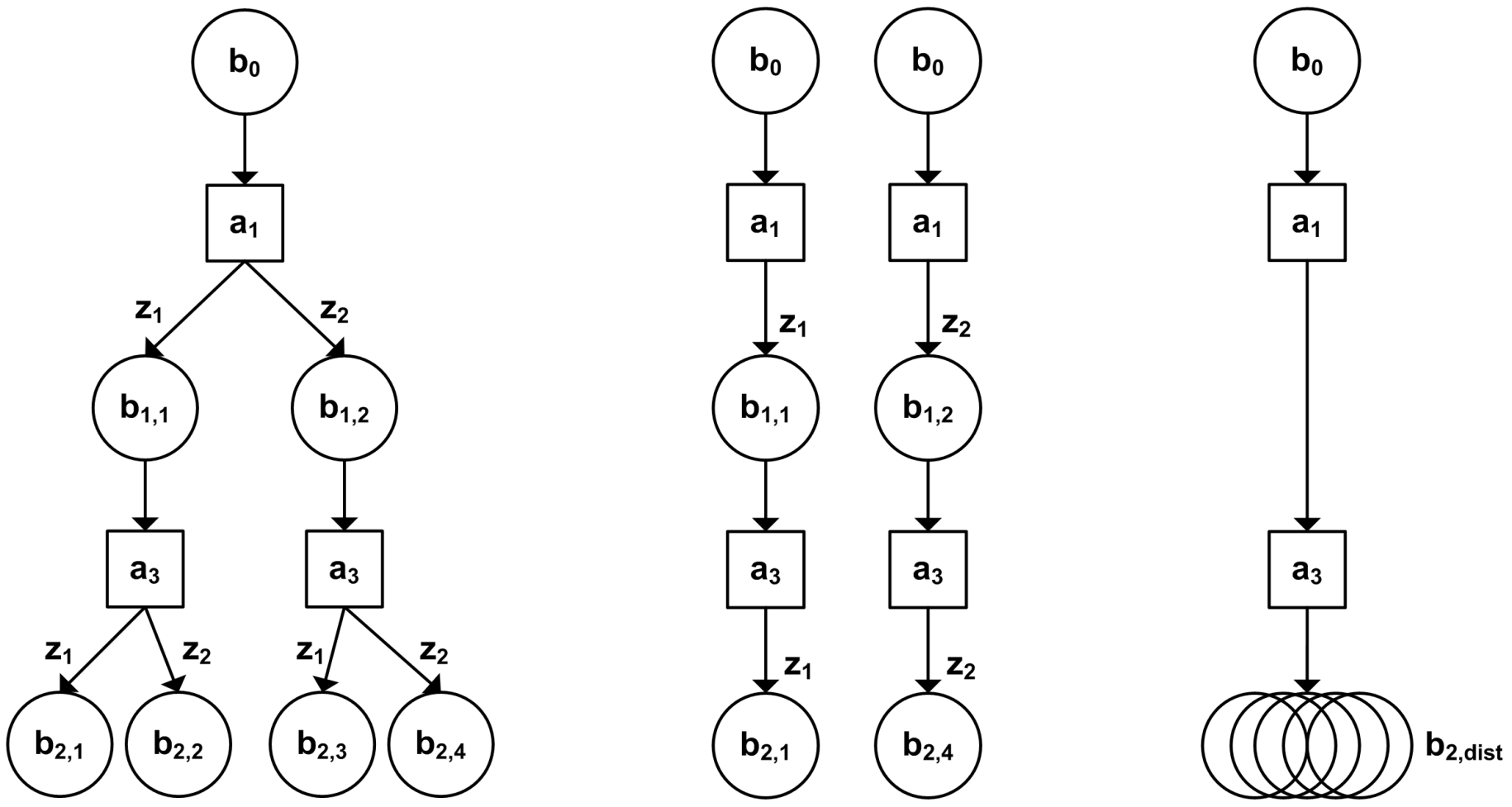

(a) Exhaustive (b) Sampled (c) Analytic

Figure 2: Three methods to represent the resulting set of beliefs after a single macro-action. (a) All possible observations are expanded. (b) A subset of possible observation trajectories are sampled. (c) Compute an analytic distribution over the posterior beliefs, which could have been generated via an exhaustive enumeration of all possible observation sequences. $b_0$ is the initial belief, while $b_{i,j}$ refers to the $j^{\text{th}}$ belief leaf node at depth $i$.

We define $b_{dist}$ as the posterior distribution over beliefs after a macro-action. We will show in the next subsection (3.1) that when the parametric form of the model is such that the belief is always Gaussian, then the distribution over posterior beliefs is itself a Gaussian over Gaussian beliefs, as illustrated in Figure 3. This property follows from the fact that all future beliefs are Gaussian. The random variables described by the distribution over posterior beliefs are therefore the means and covariances of the posterior beliefs. In this case, $b_{dist}$ consists of an expression for the distribution over belief means and an expression for the distribution over the covariances after a macro-action. We will show that the means are distributed according to a Gaussian and the covariances are a delta function over a single covariance, allowing us to represent the entire distribution over beliefs as a Gaussian distribution over beliefs means and a single belief covariance. In Section 3.2 we will further show that we can analytically compute the expected reward of the distribution over beliefs resulting from a macro-action for certain classes of reward functions. Given the ability to analytically compute a distribution over posterior beliefs, we will show in Section 5 the computational complexity of forward search is reduced to a function of the macro-action horizon $\tilde{H}$: for macro-actions of length 2 or more ($L \geq 2$) we will see that it is significantly faster to search to long horizons.

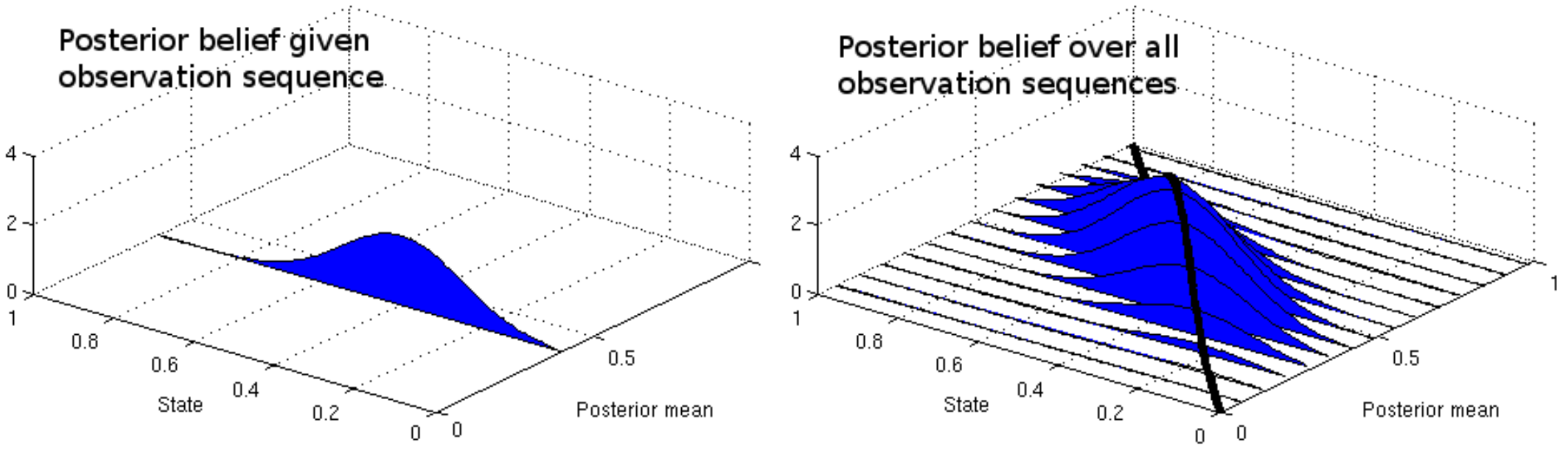

Figure 3: Distribution of posterior beliefs. a) A single Gaussian posterior belief is the result of incorporating an observation sequence. b) Over all possible observation sequences, the distribution of posterior means is a Gaussian (black line), and for each posterior mean, a Gaussian (blue curve) describes the agent's posterior belief.

### 3.1 Exact Computation of Posterior Belief Distribution

Let us assume for the moment that the agent's belief can be exactly represented as a Gaussian distribution over a continuous state space, and that the observation and transition models are both linear-Gaussian. Formally, the state transition and observation models can be represented as follows:

$$s_t = As_{t-1} + Ba_t + \varepsilon_t, \qquad \varepsilon_t \sim \mathcal{N}(0, P) \tag{5}$$

$$z_t = Cs_t + \delta, \qquad \delta \sim \mathcal{N}(0, Q) \tag{6}$$

where $A$ and $B$ are dynamics matrices, $C$ is the observation matrix, $P$ is the covariance of the Gaussian dynamics process and $Q$ is the covariance of the measurement noise.

When the state-transition and observation models are normally distributed and linear functions of the state, the Kalman filter (1960) provides a closed-form solution for the posterior belief over states, $\mathcal{N}(\mu_t, \Sigma_t)$ given a prior belief over states, $\mathcal{N}(\mu_{t-1}, \Sigma_{t-1})$,

$$\overline{\mu}_t = A\mu_{t-1} + Ba_t \qquad \mu_t = \overline{\mu}_t + K_t(z_t - C\overline{\mu}_t) \tag{7}$$

$$\overline{\Sigma}_t = A\Sigma_{t-1}A^T + P \qquad \Sigma_t = (C^TQ^{-1}C + \overline{\Sigma}_t^{-1})^{-1}, \tag{8}$$

where $\mathcal{N}(f, F)$ is a $D$-dimensional Gaussian with mean $f$ and covariance matrix $F$, $K_t = \overline{\Sigma}_tC^T(C\overline{\Sigma}_tC^T + Q)^{-1}$ is the Kalman gain and $\overline{\mu}_t$ and $\overline{\Sigma}_t$ are the mean and covariance after an action is taken but before incorporating the measurement.

Our key interest is to represent the distribution over possible beliefs that could result after taking a particular action, but receiving any of the possible observations. Note that in the current setup, all posterior beliefs are Gaussians, and can therefore be completely characterized by their mean and covariance. We now derive an expression for the distribution over the posterior belief means, under any possible observation, when the prior distribution over beliefs is simply a delta function over a single belief. We first re-express the observation model as

$$z_t \sim \mathcal{N}(Cs_t, Q) \tag{9}$$

which we can use to compute an expression for the probability of an observation given the belief mean, $p(z_t|\overline{\mu}_t)$, by marginalizing over $s_t \sim \mathcal{N}(\overline{\mu}_t, \overline{\Sigma}_t)$, as

$$p(z_t|\overline{\mu}_t) = \int p(z_t|s_t)p(s_t|\overline{\mu}_t)ds_t \tag{10}$$
$$= \mathcal{N}(C\overline{\mu}_t, C\overline{\Sigma}_t C^T + Q). \tag{11}$$

We can perform further linear transformations to obtain an expression for the distribution of posterior means, under any potential observation:

$$z_t \sim \mathcal{N}(C\overline{\mu}_t, C\overline{\Sigma}_t C^T + Q) \tag{12}$$
$$z_t - C\overline{\mu}_t \sim \mathcal{N}(0, C\overline{\Sigma}_t C^T + Q) \tag{13}$$
$$K_t(z_t - C\overline{\mu}_t) \sim \mathcal{N}(0, K_t(C\overline{\Sigma}_t C^T + Q)K_t^T) \tag{14}$$
$$\overline{\mu}_t + K_t(z_t - C\overline{\mu}_t) \sim \mathcal{N}(\overline{\mu}_t, K_t(C\overline{\Sigma}_t C^T + Q)K_t^T) \tag{15}$$
$$\mu_t \sim \mathcal{N}(\overline{\mu}_t, K_t(C\overline{\Sigma}_t C^T + Q)K_t^T) \tag{16}$$
$$\mu_t \sim \mathcal{N}(\overline{\mu}_t, \overline{\Sigma}_t C^T K_t^T) \tag{17}$$

where Equation 17 is computed by substituting the definition of the Kalman gain.

At this point, a somewhat unusual change has occurred, in that $\mu_t$, the mean of the distribution itself, is now a random variable. Without knowing the value of the particular observation that occurs after a primitive action, we cannot deterministically predict the posterior mean of the belief.[8] However, we can model the probability of any specific belief state, which effectively means that we will compute a distribution over the belief means $\mu$ and covariances $\Sigma$. Equation 17 shows that the distribution over the belief means is normally distributed about $\overline{\mu}_t$, with a covariance that depends on the prior covariance $\overline{\Sigma}_t$ and the observation model parameters. Sampling a mean from this distribution is equivalent to selecting a particular observation.

We have just presented a formula for calculating the posterior distribution over belief means after one action, and any possible observation. We now wish to show that the posterior distribution over beliefs means after a sequence of actions remains a Gaussian distribution. This will allow us to compute an analytic expression for the posterior distribution over beliefs that could result from a macro-action. We therefore require a method to iteratively use Equation 17 in order to compute the posterior distribution over beliefs for a complete macro-action and any possible observation sequence.

We first combine the process and measurement updates for a single primitive action belief update in order to get an expression for the posterior belief means in terms of the prior belief mean. We marginalize over $\overline{\mu}_t$, the posterior belief after the transition update but before the observation update, using $p(\mu_t|\mu_{t-1}) = \int p(\mu_t|\overline{\mu}_t)p(\overline{\mu}_t|\mu_{t-1})d\overline{\mu}_t$. As $\overline{\mu}_t$ is a deterministic function of $\mu_{t-1}$ (see Equation 7a), then $p(\overline{\mu}_t|\mu_{t-1})$ is simply a delta function, which means that $p(\mu_t|\mu_{t-1})$ is identical to Equation 17 after substituting $\overline{\mu}_t$ using Equation 7a:

$$p(\mu_t|\mu_{t-1}) = \mathcal{N}(A\mu_{t-1} + Ba_t, \overline{\Sigma}_t C^T K_t^T). \tag{18}$$

In a one-step belief update, the belief mean at the prior time step, $\mu_{t-1}$, is assumed to be a known value. However, for a macro-action, once the first primitive action has been taken, the posterior be-

8. Note that we will show later in this section that we can deterministically predict the posterior belief covariance. Its distribution is a Dirac delta that is independent of the specific observation received.

lief mean will depend on the received observation. In absence of the knowledge of that received observation, we will instead have a distribution over the belief means. Therefore, for the second primitive action in the macro-action, the prior belief is now given as a Gaussian $\mu_{t-1} \sim \mathcal{N}(m_{t-1}, \Sigma^{\mu}_{t-1})$ where $m_{t-1}$ and $\Sigma^{\mu}_{t-1}$ are random variables. In order to compute the probability distribution over $\mu_t$, we must integrate over this distribution of prior belief means $\mu_{t-1}$:

$$p(\mu_t|m_{t-1}, \Sigma^{\mu}_{t-1}) \quad = \quad \int_{\mu_{t-1}} p(\mu_t|\mu_{t-1})p(\mu_{t-1}|m_{t-1}, \Sigma^{\mu}_{t-1})d\mu_{t-1}. \tag{19}$$

Since both terms inside the integral are Gaussian distributions, we can analytically combine these two Gaussians, one of which is independent of $\mu_{t-1}$ and one of which is dependent on $\mu_{t-1}$. Integrating over $\mu_{t-1}$, as we had done in Equations 9-11, we find that the mean of the posterior belief means is conveniently still a Gaussian distribution over a function of the prior mean of the belief means and covariance:

$$\mu_t \quad \sim \quad \mathcal{N}(Am_{t-1} + Ba_t, A\Sigma^{\mu}_{t-1}A^T + \overline{\Sigma}_t C^T K_t^T) \tag{20}$$

or

$$\mu_t \quad \sim \quad \mathcal{N}(m_t, \Sigma^{\mu}_t) \tag{21}$$

where $m_t = Am_{t-1} + Ba_t$ and $\Sigma^{\mu}_t = A\Sigma^{\mu}_{t-1}A^T + \overline{\Sigma}_t C^T K_t^T$. Equation 20 can now be used to predict the posterior mean distribution after a multi-step action sequence. Assuming that the agent is currently at time $t$ and has a particular prior mean $\mu_t$ (which we can also express as a Gaussian with zero covariance, $\mathcal{N}(\mu_t, 0)$), the posterior mean after an action sequence of $D$ time steps is distributed as follows:

$$\mu_{t+D} \sim \mathcal{N}(m_{t+D}, \Sigma^{\mu}_{t:t+D}) \tag{22}$$

where

$$\begin{aligned} m_{t+D} &= f(\mu_{t-1}, A, B, a_{t+1:t+D}) && (23)\\ &= A\, m_{t+D-1} + B\, a_{t+D} && (24)\\ &= A^D m_t + \sum_{i=1}^{D} A^{D-i} B a_{t+i}, && (25)\end{aligned}$$

and

$$\Sigma^{\mu}_{t:t+D} = \sum_{i=t}^{t+D} A^{t+D-i}\overline{\Sigma}_i C_i^T D_i^T (A^{t+D-i})^T. \tag{26}$$

Note that $m_{t+D}$ does not depend on observations; it gives the mean of the distribution of beliefs that might result from the received observations. $m_{t+D}$ is dependent only on the state-transition model parameters and can be calculated via a recursive update along the action sequence.

We now consider the covariance of the posterior beliefs that may result after taking a macro-action. Recall that for a single belief, the posterior covariance after taking a primitive action and receiving a particular observation can be calculated using Equation 8. Note that this formula is independent of the actual received observation $z_t$, and the prior $\mu_{t-1}$ or posterior mean $\mu_t$. Formally, this property exists because the Fisher information associated with the observation model is independent of the specific observations. Therefore, the posterior covariance after any observation sequence of known length can be calculated in closed form given the prior covariance, without needing to know the observations received along the way.

We can now specify the form of $b_{dist}$, the posterior distribution over beliefs after a macro-action:

$$b_{dist}(\mu_{t+T}, \Sigma) = \mathcal{N}(f(\mu_{t-1}, A, B, a_{t:t+T}), \Sigma^{\mu}_{t:T}) \cdot \delta(\Sigma, \Sigma') \tag{27}$$

where $b_{dist}(\mu_{t+T}, \Sigma)$ is the probability of arriving in posterior belief $b = \mathcal{N}(\mu_{t+T}, \Sigma)$ after taking a particular macro-action, Equation 22 defines the distribution over belief means, and $\Sigma'$ is computed by iteratively applying Equation 8. This expression shows that for problems with linear-Gaussian state-transition and observation models, we can exactly calculate the distribution of posterior beliefs associated with a macro-action.

### 3.2 Calculating the Expected Reward

The prior section outlined a procedure for calculating the posterior set of beliefs after a macro-action. The reason to compute this distribution is in turn to be able to calculate the expected reward of each macro-action, which will be used to compute the best action for the current belief.

To calculate the expected reward of a macro-action, we start by considering the expected reward of starting in a particular belief state $b_0$ and executing a $L$-length macro-action $\tilde{a}$ consisting of actions $a_1, a_2, \ldots, a_L$. This may be expressed as

$$r(b_0, \tilde{a}_{1:L}) = r(b_0, a_1) + \gamma \int_{z_1} p(z_1|b_0, a) Q(b^{a_1,z_1}, \tilde{a}_{2:L}) \tag{28}$$

where we have used $b^{a_1,z_1}$ to represent the updated belief after taking action $a_1$ and receiving observation $z_1$ from $b_0$, $\tilde{a}_{2:L}$ to represent the macro-action consisting of the second through $L$-th primitive actions of the macro-action $\tilde{a}$, and $Q(b^{a_1,z_1}, \tilde{a}_{2:L})$ to represent the future expected reward of taking the remaining actions from belief $b^{a_1,z_1}$. Recursively expanding the second term in Equation 28 we obtain the following expression

$$\begin{aligned} r(b_0, \tilde{a}_{1:L}) \quad &= \quad r(b_0, a_1) + \gamma \int_{z_1} p(z_1|b_0, a_1) r(b^{a_1,z_1}, a_2) + \\ &\quad \gamma^2 \int_{z_1,z_2} p(z_1|b_0, a_1) p(z_2|b^{a_1,z_1}, a_2) r(b^{a_1,z_1,a_2,z_2}, a_3) + \cdots \quad (29) \\ &\quad \gamma^{L-1} \int_{z_1,\ldots,z_L} \left[ \prod_{i=1}^{L-1} p(z_i|b^{a_1,z_1,\ldots,a_{i-1},z_{i-1}}, a_i) \right] r(b^{a_1,\ldots a_{L-1},z_{L-1}}, a_L). \quad (30) \end{aligned}$$

The first term in Equation 29 represents the expected reward from taking the first primitive action in the macro-action from the initial belief state. The remaining terms each represent the expected reward at the $i$-th primitive action of the macro-action, where the expectation is taken over all possible $i-1$ length sequences of observations that could have been received up to that point (as well as the standard integration over the state space). From Equation 27 we have a closed form expression for the distribution over belief states possible after a sequence of primitive actions. We can use this to re-express Equation 29 as a function of the distributions over beliefs:

$$r(b_0, \tilde{a}_{1:L}) = r(b_0, a_1) + \sum_{i=2}^{L} \gamma^{i-1} r(b^{i-1}_{dist}, a_i) \tag{31}$$

where $b_{dist}^{i-1}$ is used to represent the posterior distribution over beliefs that results after taking the first $i-1$ primitive actions in macro action $\tilde{a}$. Slightly abusing notation, $r(b_{dist}, a_i)$ represents the expected reward for taking action $a_i$ given the posterior distribution over beliefs $b_{dist}$, and is expressed as

$$r(b_{dist}, a_i) = \int_b \int_s b(s) b_{dist}(b) r(s, a_i) ds db. \tag{32}$$

Combining Equations 31 and 32, we can see that the expected reward of a macro-action can be calculated from the sum of the expected reward of taking a primitive action from the posterior distribution of beliefs at each step along the macro-action.

Recall from the prior section that the posterior distribution over beliefs can be factored into a Gaussian distribution over the belief means $\mu$ (Equation 22), and a Dirac delta distribution over the belief covariances $\Sigma$ (since all beliefs will have identical covariances):

$$b_{dist}(\mu, \Sigma) = \mathcal{N}(\mu | m_a, \Sigma_a^{\mu}) \delta(\Sigma, \Sigma_a) \tag{33}$$

where $m_a$ is the mean of the belief means after primitive action $a$, $\Sigma_a^{\mu}$ is the covariance of the belief means after primitive action $a$, and $\Sigma_a$ is the covariance of a belief state after primitive action $a$.

As the belief state itself is a Gaussian,

$$b(s) = \mathcal{N}(s | \mu, \Sigma), \tag{34}$$

we can re-express the reward as

$$\begin{aligned} r(b_{dist}, a) &= \int_s \int_{\mu,\Sigma} r(s, a) \mathcal{N}(s|\mu, \Sigma) \mathcal{N}(\mu | m_a, \Sigma_a^{\mu}) \delta(\Sigma, \Sigma_a) ds d\mu d\Sigma & (35) \\ &= \int_s \int_{\mu} r(s, a) \mathcal{N}(s|\mu, \Sigma_a) \mathcal{N}(\mu | m_a, \Sigma_a^{\mu}) d\mu ds, & (36) \end{aligned}$$

where the second line follows due to the Dirac delta distribution on the belief covariances. Expanding out the formula for $\mathcal{N}(s|\mu, \Sigma)$ we see it is identical to the formula for $\mathcal{N}(\mu|s, \Sigma)$:

$$\begin{aligned} \mathcal{N}(s|\mu, \Sigma) &= \frac{1}{\sqrt{2\pi} |\Sigma|^{N_d/2}} exp(-\frac{1}{2}(s-\mu)\Sigma^{-1}(s-\mu)^T) & (37) \\ &= \frac{1}{\sqrt{2\pi} |\Sigma|^{N_d/2}} exp(-\frac{1}{2}(\mu-s)\Sigma^{-1}(\mu-s)^T) & (38) \\ &= \mathcal{N}(\mu|s, \Sigma). & (39) \end{aligned}$$

Therefore, we can substitute the equivalent expression to yield

$$r(b_{dist}, a) = \int_s \int_{\mu} r(s, a) \mathcal{N}(\mu|s, \Sigma_a) \mathcal{N}(\mu | m_a, \Sigma_a^{\mu}) d\mu ds. \tag{40}$$

Completing the square in the exponent, we re-express the product of the above two Gaussians as

$$r(b_{dist}, a) = \int_s \int_{\mu} r(s, a) \mathcal{N}(s|m_a, \Sigma_a + \Sigma_a^{\mu}) \mathcal{N}(\mu|\hat{c}, \hat{C}) d\mu ds, \tag{41}$$

where $\hat{C} = (\Sigma_a^{-1} + (\Sigma_a^\mu)^{-1})^{-1}$ and $\hat{c} = \hat{C}(m_a(\Sigma_a^\mu)^{-1} + \mu\Sigma_a^{-1})$. We then integrate over $\mu$ to get

$$r(b_{dist}, a) \quad = \quad \int_s r(s,a)\mathcal{N}(s|m_a, \Sigma_a + \Sigma_a^\mu)ds. \tag{42}$$

If the reward model itself is a weighted sum of $N_r$ Gaussians,

$$r(s,a) = \sum_{j=1}^{N_r} w_j \mathcal{N}(s|\zeta_j, \Upsilon_j), \tag{43}$$

then the integral in Equation 42 can be evaluated in closed form as

$$r(b_{dist}, a) \quad = \quad \int_s \sum_{j=1}^{N_r} w_j \mathcal{N}(s|\zeta_j, \Upsilon_j)\mathcal{N}(s|m_a, \Sigma_a + \Sigma_a^\mu)ds \tag{44}$$

$$= \quad \sum_{j=1}^{N_r} w_j \mathcal{N}(\zeta_j|m_a, \Upsilon_j + \Sigma_a + \Sigma_a^\mu)\int_s \mathcal{N}(s|c_1, C_1), \tag{45}$$

where we have again completed the square in the exponent, and defined new constants $C_1 = (\Upsilon_j^{-1} + (\Sigma_a + \Sigma_a^\mu)^{-1})^{-1}$ and $c_1 = C_1(\zeta_j\Upsilon_j^{-1} + m_a(\Sigma_a + \Sigma_a^\mu)^{-1})$. Integrating we obtain an analytic expression for the expected reward of a primitive action under a distribution of beliefs:

$$r(b_{dist}, a) \quad = \quad \sum_{j=1}^{N_r} w_j \mathcal{N}(\zeta_j|m_a, \Upsilon_j + \Sigma_a + \Sigma_a^\mu). \tag{46}$$

A similar closed-form expression is available if the reward model is a polynomial function of the state,

$$r(s,a) = \sum_{j=1}^{N_r} w_j s^j, \tag{47}$$

instead of a weighted sum of Gaussians. Substituting Equation 47 into Equation 42 yields

$$r(b_{dist}, a) \quad = \quad \int_s \sum_{j=1}^{N_r} w_j s^j \mathcal{N}(s|m_a, \Sigma_a + \Sigma_a^\mu)ds$$

$$= \quad \sum_{j=1}^{N_r} w_j \int_s s^j \mathcal{N}(s|m_a, \Sigma_a + \Sigma_a^\mu)ds. \tag{48}$$

Therefore, evaluating the expected reward involves calculating the first $N_r$ moments of a Gaussian distribution. Each of these moments is an analytic expression of the Gaussian mean and covariance.[9] So, for reward models that are either a weighted sum of Gaussians, or which are polynomial functions of the state space, the expected reward of a macro-action (Equation 28) can be computed analytically.

For other arbitrary reward models it may not be possible to analytically compute the expected reward of taking a primitive action in a particular distribution over beliefs. In such cases, we can approximate the expectation in Equation 42 by sampling.

9. The Gaussian distribution is completely described by its first two moments; all higher order moments are simply functions of the first two moments.

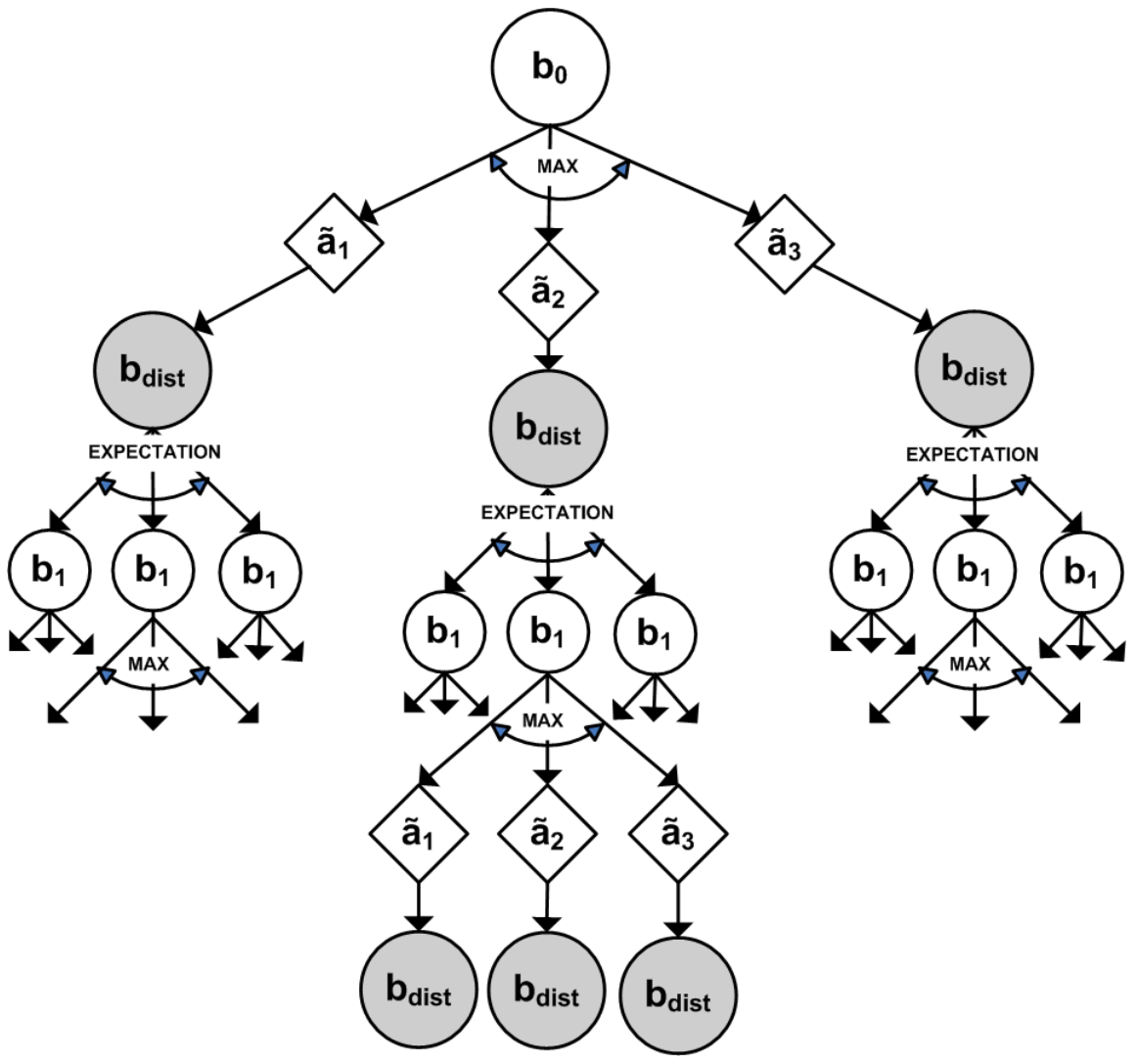

Figure 4: In PBD, individual beliefs $b$ are sampled from the posterior distribution over beliefs $b_{dist}$, implicitly sampling a particular observation trajectory. Then the best macro-action is selected for each sampled posterior belief. A sum is taken over all the sampled beliefs, again corresponding to a sum over the implicitly sampled observation sequences. Here, $b_i$ refers to beliefs at macro-action depth $i$.

### 3.3 Branching on Posterior Beliefs

So far we have discussed how to compute the posterior distribution over beliefs that can arise after executing a single macro-action, and how to compute the expected reward associated with that distribution. But during planning we wish to compute the value of not taking just a single macro-action, but sequences of macro-actions. This allows us to consider scenarios much further in the future, which can be useful in selecting the best action to take for the current belief. For example, consider a large office space domain where a robot is trying to navigate to a goal location, and macro-actions are to go to the end of a hallway and turn left or right. Assuming the robot starts far from the goal location, a series of macro-actions will most likely be needed in order to reach the goal, and therefore it will be important during forward search to consider a search horizon of multiple macro-actions.

However, when constructing the forward search tree, it is not immediately clear how to evaluate each branch in the three at the end of each macro-action. We have a closed form expression for the posterior distribution over beliefs at the end of the macro-action. This posterior set represents the distribution of beliefs possible given *all* possible observation sequences that could be received during the macro-action's execution. However, different individual posterior beliefs, or different subsets of the posterior belief distribution, may be associated with different best subsequent macro-actions in the tree, because different individual posterior beliefs are implicitly the result of receiving a different set of observations during the macro-action execution and may reveal important information about the environment that result in different best subsequent macro-actions. Though the motivation behind macro-actions is that it is reasonable to act in an open-loop fashion for a limited

**Algorithm 1** Forward Search with Macro-Actions

**Require:** Initial belief $b_0$, Discount factor $\gamma$, Macro-action search depth $\tilde{H}$, Sampling number $N_s$

$t \leftarrow 0$
**loop**
  Compute set of macro-actions $\tilde{\mathcal{A}}$
  **for** each macro-action $\tilde{a}_i \in \tilde{\mathcal{A}}$ **do**
    $Q(b_t, \tilde{a}_i) = \textsc{Expand}(\tilde{a}_i, b_t, \gamma, \tilde{H}, N_s)$ {See Algorithm 2}
  **end for**
  Execute first action $a_1$ of $\tilde{a} = \operatorname{argmax}_{\tilde{a}} Q(b_t, \tilde{a})$
  Obtain new observation $z_t$ and reward $r_t$
  $b_{t+1} = \tau(b_t, a_t, z_t)$
  $t \leftarrow t + 1$
**end loop**

time period, the received observation sequence does provide information about the underlying belief that is likely to be useful for selecting future macro-actions.

Since we do not know in advance which subsets of posterior beliefs are associated with the same best subsequent macro-action, we instead sample from the posterior belief distribution, and then evaluate future macro-actions for each of these samples (see Figure 4 for an illustration). Sampling a posterior belief is equivalent to implicitly sampling an observation sequence for the planned macro-action, without having to actually perform belief updates along the action-observation trajectory. Note that the potential space of observation sequences grows exponentially with the macro-action length. As the posterior distribution over beliefs is a Gaussian, its properties can be completely described by its mean and covariance, which means that the posterior distribution over beliefs will typically be of much lower dimension than the observation sequence space. Experimentally we will see much better performance sampling from the posterior belief distribution than from sampling from the space of observation sequences. The sampled beliefs essentially form a non-parametric, particle estimate of the posterior distribution of beliefs that is present after taking the macro-action. As the number of samples $N_s$ goes to infinity, the sampled distribution will become an arbitrarily good approximation of the full posterior distribution of beliefs. As the covariance is a Dirac delta distribution, sampling is needed only for the posterior mean distribution, generating posterior belief samples by associating each posterior mean sample with the posterior covariance $\Sigma_{t+T}$.

### 3.4 The PBD Algorithm Summary

We are now ready to present our PBD forward search algorithm (Algorithm 1). Given the current belief, we select an action by constructing a macro-action forward search tree. Placing the current belief at the root, we expand each possible macro-action (Algorithm 2), computing the expected reward and the resulting posterior set of beliefs. We then sample a fixed number of posterior beliefs. Forward search then proceeds from each of these sampled beliefs. We repeat this process out to a fixed horizon depth and then select an action for the current belief by estimating its value, starting from the search leaf nodes. After executing this action, an observation is received, and the new belief state is computed. The whole process then repeats for this new belief state. Note that PBD will only ever select actions that are the first action of a macro-action. If all primitive actions are to

**Algorithm 2** EXPAND – Expand Macro-actions via PBD

**Input:** Macro-action $\tilde{a}$, Belief state $b_t$, Discount factor $\gamma$, Macro-action search depth $\tilde{H}$, No. posterior belief samples per macro-action $N_s$
**if** $\tilde{H} = 0$ **then**
  return 0
**else** $\{$Expand Macro-action $\tilde{a}$=$\{a_1, \ldots, a_L\}\}$
  $R_{\tilde{a}} = 0$
  $b_{dist} = b_t$
  **for** $j = 1$ to $L$ **do**
    $R_{\tilde{a}} = R_{\tilde{a}} + \gamma * r(b_{dist}, a_j)$
    Update the posterior distribution of beliefs $b_{dist}$
  **end for**
  **for** $i = 1$ to $N_s$ **do**
    Sample posterior mean $n_i$ according to $\mathcal{N}(m_{t+T}, \Sigma^{\mu}_{t+T})$
    $b_i \leftarrow \mathcal{N}(n_i, \Sigma_{t+T})$
    Generate next set of macro-actions $\tilde{\mathcal{A}}^{next}$
    **for** $\tilde{a}^{next}_i \in \tilde{\mathcal{A}}^{next}$ **do**
      $Q(b_i, \tilde{a}^{next}_i)$ = EXPAND($\tilde{a}^{next}_i$,$b_i$,$\gamma$,$\tilde{H} - 1$,$N_s$)
    **end for**
    $V = R_{\tilde{a}} + \frac{1}{N_s}\gamma^L \max_{\tilde{a}^{next}_i} Q(b_i, \tilde{a}^{next}_i))$
  **end for**
  return $V$
**end if**

be considered, the number of macro-actions that are evaluated for the root belief at every timestep must be at least the same as the size of the primitive action space, and each primitive action must be the first action of at least one macro-action.

## 4. Approximate Computation of Posterior Belief Distributions

The PBD algorithm described so far assumes that the transition and observation functions are linear functions of the state with Gaussian noise. When these functions are non-linear, the traditional Kalman filter model no longer provides an exact belief update, and for the PBD algorithm, the distribution of posterior beliefs cannot be calculated exactly. In this section we briefly describe an extension to the PBD algorithm to handle a wider class of observation models, namely parametric models that are members of the exponential family of distributions (Barndorff-Nielsen, 1979). For non-linear transition models, there exist techniques such as the extended Kalman filter to approximate the posterior with a Gaussian; however, we do not formally consider incorporating such techniques into our PBD algorithm here.

We choose to consider exponential family observation models since this family includes a wide array of distributions, such as Gaussian, Bernoulli, and Poisson distributions, and has certain appealing mathematical properties. In particular, we leverage work by West, Harrison and Migon (1985) who constructed linear-Gaussian models that approximate the non-Gaussian exponential family observation model in the neighborhood of the conditional mode, $s_t|z_t$. They then used the approximate linear-Gaussian observation mode in a traditional Kalman filter, to maintain a closed-form Gaussian representation of the posterior belief, creating an exponential family Kalman Filter (efKF). For completeness we include West et al.'s derivation of the filter in Appendix A, and we present the main equations here.

Constructing the approximate linear-Gaussian observation model requires computation of the first two moments of the distribution and the linearization around the mean estimate at every time step. An exponential family observation model can be represented as follows,

$$p(z_t|\theta_t) = \exp(z_t^T\theta_t - \beta_t(\theta_t) + \kappa_t(z_t)), \qquad \theta_t = W(s_t) \tag{49}$$

where $s_t$ is the hidden state of the system, $\theta_t$ and $\beta_t(\theta_t)$ are the canonical parameter and normalization factor of the distribution, and $W(.)$ maps the states to canonical parameter values. $W(.)$ is also known as the canonical link function, and depends on the particular member of the exponential family.

The first two moments of the distribution (West et al., 1985) are

$$E(z_t|\theta_t) = \dot{\beta}_t = \frac{\partial\beta_t(\theta_t)}{\partial\theta_t}\Big|_{\theta_t=W(\overline{\mu}_t)} \quad Var(z_t|\theta_t) = \ddot{\beta}_t = \frac{\partial^2\beta_t(\theta_t)}{\partial\theta_t\partial\theta_t^T}\Big|_{\theta_t=W(\overline{\mu}_t)} \tag{50}$$

where $\dot{\beta}_t$ and $\ddot{\beta}_t$ are the derivatives of the exponential family distribution's normalization factor, both linearized about $\overline{\theta}_t = W(\overline{\mu}_t)$.

Given an action-observation sequence, the posterior mean of the agent's belief in the efKF can then be updated according to

$$\overline{\mu}_t = A\mu_{t-1} + Ba_t \qquad \mu_t = \overline{\mu}_t + \tilde{K}_t(\tilde{z}_t - W(\overline{\mu}_t)), \tag{51}$$

$$\overline{\Sigma}_t = A\Sigma_{t-1}A^T + P \qquad \Sigma_t = (\overline{\Sigma}_t^{-1} + Y_t^T\ddot{\beta}_tY_t)^{-1}, \tag{52}$$

where $\tilde{K}_t = \overline{\Sigma}_tY_t(Y_t\overline{\Sigma}_tY_t^T + \ddot{\beta}_t^{-1})^{-1}$ is the efKF Kalman gain, and $\tilde{z}_t = \overline{\theta}_t - \ddot{\beta}_t^{-1} \cdot (\dot{\beta}_t - z_t)$ is the projection of the observation onto the parameter space of the exponential family observation model. $Y_t = \frac{\partial\theta_t}{\partial s_t}\big|_{s_t=\overline{\mu}_t}$ is the gradient of the exponential family distribution's canonical parameter, linearized about $\overline{\mu}_t$.

We can now incorporate these results to compute a modified form for the posterior belief mean and covariance distributions, which were represented by Equations 8 and 22 when the observation model was linear Gaussian. Now, for exponential family observation models, the posterior belief covariance comes from Equation 52. The expression for the distribution of the posterior means can be modified based on the efKF equations:

$$\mu_{t+T} \sim \mathcal{N}(f(\mu_{t-1}, A_{t:t+T}, B_{t:t+T}, a_{t:t+T}), \sum_{i=t}^{t+T} \overline{\Sigma}_iY_i^T\tilde{K}_i^T). \tag{53}$$

It is worth noting that in contrast to our prior expressions for the posterior belief distribution (Equations 8 and 22), which are exact and completely independent of the received observations, Equations 52 and 53 are no longer independent of the observations obtained because the observation model parameters are linearized about the prior mean $\overline{\mu}_t$. Hence while the parameters are independent of the observation that will be obtained for a macro-action sequence of length 1, for a longer macro-action, the observation model parameters depend on the prior observations obtained. We approximate this update by linearizing about the mean of the prior mean distribution $m_t$ at each step along the action sequence, rather than the true prior belief mean $\mu_t$. We will shortly see that we still obtain good experimental results using this approximation.

An alternate popular approach for non-Gaussian systems is to use a particle filter to represent the system state. However, in high dimensional, continuous environments similar to the ones considered in this paper, particle filters often suffer from particle depletion, or require a very large number of particles to accurately capture the posterior. The costs of belief updating and expected reward calculations scale with the number of particles. In contrast, our approximate PBD computation has the same computational complexity as our exact PBD computation, which we will demonstrate in later sections to scale polynomially with the number of state dimensions.

This approximate method for computing the posterior distribution over beliefs can be used as a substitute for exactly calculating the posterior distribution over beliefs in the PBD algorithm.

## 5. Analysis

Here we provide a formal analysis of the accuracy and computational complexity of our PBD algorithm. Throughout this section we assume belief states can be represented exactly as Gaussian distributions: in other words, we assume a linear-Gaussian system. In the following sections we will demonstrate experimentally that the PBD algorithm is useful in a wider variety of problems using an EKF or the efKF described in Section 4, but incorporating the error of these approximate filtering techniques into an analysis of the algorithm is a topic for future research.

### 5.1 Performance

PBD selects actions by performing a limited-horizon forward search using a restricted policy space induced by the macro-actions. However, during execution, only the first step of the macro-action is taken. After an observation is received, the belief state is updated, and then planning is repeated from the resulting belief. By only taking the first primitive action, the system may take sequences of actions that do not correspond to any of the known macro-actions, effectively expanding the considered policy space. As a result, the performance will be at least as good as actually executing the entire macro-action. However, it would be useful to determine if any claims can be made about the belief-action values calculated as part of the PBD algorithm. Obviously, the received rewards of the executed policy will always be less than or equal to the optimal policy's rewards, since the policy space considered during planning is smaller than the full policy space. However, the values *calculated* by the PBD algorithm are only approximate values due to the approximations (such as sampling a subset of the posterior beliefs) made during the computation process. We now prove that for linear-Gaussian systems, the values computed by PBD, minus an additional epsilon term due to the approximations incurred by sampling a subset of the posterior beliefs after each macro-action, are probabilistically guaranteed to be a lower bound on the true optimal values. For the purpose of this analysis we will assume that all rewards are scaled to lie between 0 and 1. $M$ is the maximum number of macro-actions.

**Theorem 5.1** *Given a linear-Gaussian system, an initial belief* $b$*, and any* $\delta > 0$*, and for any reward model which is either a weighted sum of Gaussians, or a polynomial function, the following* *lower bound on the optimal value of* $b$ *holds*

$$V_{PBD}(b) - \epsilon_{\tilde{H}} \leq V^*(b)$$

*with probability at least* $1 - \delta$*, where* $\epsilon_{\tilde{H}} = \gamma^H V_{max} + \frac{1}{1-\gamma}(\sqrt{\frac{V_{max}^2}{N_s}\log(\frac{(MN_s)^{\tilde{H}}}{\delta})})$*,* $V_{max}$ *is a bound on the maximum value*[10]*, and* $V_{PBD}(b)$ *is the best value computed for* $b$ *by the PBD planning algorithm.*

**Proof** First recall in the PBD algorithm that after each macro-action, a subset of the possible posterior beliefs are sampled from the posterior belief distribution, before the tree is further expanded. Note that this is equivalent to implicitly sampling a subset of the observation trajectories that might have been received during that macro-action: each sampled posterior belief corresponds to the belief that would result by following the macro-action and receiving a particular sequence of observations. Consider an alternate variant of a macro-action forward search in which observation sequences are exhaustively enumerated[11]: that is, for each macro-action of length $L$, all $|Z|^L$ possible observation sequences are expanded. In this case, the forward search tree constructed is precisely a subset of a full POMDP forward search tree, since the macro-actions mean that only a subset of actions are expanded. Therefore, the computed values of this alternate algorithm are directly a lower bound on the optimal finite-horizon value, since the policy space considered is a strict subset of the full optimal finite-horizon policy space.

However, for computational reasons, at each macro-action tree node, only a subset of observation sequences are sampled, and the results are averaged across the observation sequences. As observation sequences that happen to lead to higher rewards may be, by chance, disproportionately sampled, the resulting $V_{PBD}$ value could be an upper bound to the true optimal value. However, we can now probabilistically bound this error induced by observation sampling,

Prior work by Kearns, Mansour and Ng (2002) proved bounds on the MDP state values computed using a sampled-states forward search given certain constraints on the number of samples, and the horizon of the forward search. McAllester and Singh (1999) extended these ideas to POMDPs, showing that similar bounds on the calculated values of a POMDP belief state could be computed if a sufficient number of observations were sampled, and forward search was computed out to a sufficiently large horizon. These results can be applied with little modification to our PBD algorithm. Essentially we can consider a new meta-POMDP in which the only available actions are macro-actions, and observations are sequences of primitive observations. Since we can compute the expected reward of macro-actions analytically (due to the assumed form of the reward model), the only errors in evaluating the root belief node values for a macro-action policy come from limited sampling of the observation trajectories, and performing a finite horizon lookahead. The prior results of McAllester and Singh directly apply to our meta-POMDP, and therefore, the values computed by PBD.

To obtain our final result, we depart slightly from the presentation of Kearns, Mansour and Ng who sought to compute the number of samples required, and the horizon required, to ensure the resulting root state-action values were within a specified $\epsilon$ bound of the true value. In contrast, we seek to compute the resulting error from an input number of samples $N_s$ and fixed horizon $\tilde{H}$.

10. The maximum value can be trivially upper bounded by $\max_{s,a} r(s,a)/(1-\gamma)$.
11. This is possible only if there are a finite number of observations.

In the proof of Kearns, Mansour and Ng, they show that the error between the calculated $\tilde{H}$-horizon state-action value $Q_{\tilde{H}}(b,a)$ and the true infinite-horizon policy value $Q(b,a)$ is

$$|Q_{\tilde{H}}(b,a) - Q(b,a)| \leq \gamma^{\tilde{H}} V_{max} + \frac{\epsilon}{1-\gamma} \tag{54}$$

with probability at least $1-\delta$ if

$$\delta \geq (MN_s)^{\tilde{H}} exp(-\epsilon^2 N_s / V_{max}^2). \tag{55}$$

We can solve Equation 55 for $\epsilon$, to yield

$$\epsilon \leq \sqrt{\frac{V_{max}^2}{N_s} \log\left(\frac{(MN_s)^{\tilde{H}}}{\delta}\right)}. \tag{56}$$

Substituting Equation 56 into Equation 54 and re-arranging yields the desired result. □

If the reward of a macro-action cannot be analytically computed, we can approximate its value by sampling $N_r$ samples at each primitive action along the length-$L$ macro-action. For an input $\delta'$ we can compute a probabilistic bound on the resulting error of the approximate value at each primitive action using Chernoff's bound. Using the union bound, the probability that the true error will exceed this threshold at any primitive action along the macro-action is no more than $L\delta'$, and the resulting error is at most the sum of the error at each primitive action. This error (and probability of error) can be easily incorporated to extend Theorem 1 to the case of generic reward models.

Note that Theorem 1 only states that with high probability that $V_{PBD} - \epsilon_{\tilde{H}}$ is a lower bound on the optimal value: it does not provide a tight bound on how close the computed $V_{PBD}$ is to the optimal value. To state this in an alternate way, $\epsilon_{\tilde{H}}$ provides a bound on the error introduced by sampling observation sequences, but PBD still is designed to only search over a limited policy space, that defined by the macro-actions chosen and used in the forward search. Therefore in general the computed values, even when a large number of observation sequences are sampled, may be substantially less than the value under the optimal policy.

## 5.2 Computational Complexity

One of the central contributions of our work is providing an efficient macro-action forward search algorithm that can scale to long horizons and large problems. We now analyze the computational complexity of our approach. The computational cost will be a function of two operations: computing the posterior distribution over beliefs, and computing the expected reward of a distribution over beliefs. As we will shortly see, the computational complexity of these operations is a polynomial function of the state space dimension.[12] This low order relationship is possible due to the particular parametric representation employed for the posterior distribution over beliefs: representing the posterior distribution over beliefs as a Gaussian requires a number of parameters that scales only quadratically with the number of state dimensions.[13] PBD is therefore able to scale to large domains. Our computational complexity results are summarized in Table 1. Throughout this analysis

12. If there are multiple independent state variables, or factors, the complexity increases linearly with the number of independent factors.
13. To represent a Gaussian in $X$ dimensions requires an $X$-dimensional vector to specify the mean, and $\mathcal{O}(X^2)$ parameters to specify the covariance.

we presume that the macro-actions themselves were selected or computed in advance; in general, the cost of computing domain-relevant macro-actions will depend on the particular domain, and we do not here analyze the possible additional computational cost incurred during macro-action construction.

#### 5.2.1 COMPLEXITY OF GAUSSIAN BELIEF UPDATING FOR A LENGTH L MACRO-ACTION

The computation for the posterior distribution over beliefs resulting from a macro-action was presented in Equation 53, and consists of a set of matrix multiplications and inversions. Matrix multiplication is an $\mathcal{O}(D^2)$ computation, where $D$ is the state space dimension. Matrix inversion can be done in $\mathcal{O}(D^3)$ time. Therefore the computational cost of performing a single update of the posterior over belief states is an $\mathcal{O}(D^3)$ operation. This update must be performed for each primitive action in a length-$L$ macro-action $\tilde{a}$, resulting in a computational cost of

$$\mathcal{O}(LD^3) \tag{57}$$

for a single macro-action.

In Section 4 we presented a set of equations (Equations 50- 53) that we use to approximately compute the posterior distribution over beliefs when the observation model is not Gaussian, but is an exponential family. These equations again consist of a set of matrix multiplications, and the cost of a single update, and cost of updating over a length-$L$ macro-action will again be $\mathcal{O}(D^3)$ and $\mathcal{O}(LD^3)$, respectively.[14]

#### 5.2.2 COMPLEXITY OF ANALYTICALLY COMPUTING THE EXPECTED REWARD OF A LENGTH L MACRO-ACTION

The second component of the computational cost comes when we evaluate the expected reward of a macro-action. If the reward is a weighted sum of $N_r$ Gaussians, as specified by Equation 43, this operation involves evaluating the value of $N_r L$ Gaussians at particular fixed points. Evaluating a $D$-dimensional Gaussian at a single point is an $\mathcal{O}(D^3)$ operation, due to the inverse covariance that must be computed. The cost for performing this operation $N_r L$ times is simply $\mathcal{O}(N_r LD^3)$. Therefore the total cost for evaluating the expected reward of a macro-action when the reward model is a weighted sum of $N_r$ Gaussians is:

$$\mathcal{O}(LD^3(N_r + 1)). \tag{58}$$

If instead the reward model is a $N_r$-th degree polynomial function of the state, then the expected reward calculation consists of the cost of calculating the $N_r$-moments of a $D$-dimensional Gaussian distribution (Equation 48). Assume without loss of generality that we are computing the $N_r$-th central moment of a $D$-dimensional Gaussian: a non-central moment can always be converted into a central moment by adding and subtracting a mean term. Let the $N_r$-th central moment denote moments of the form $E[(s_1 - E[s_1])^2(s_2 - E[s_2]) \ldots (s_D - E[s_D])]$ or $E[(s_2 - E[s_2])^{N_r}]$, and $\sigma_{ij}$ denote the $ij$-th entry of the covariance matrix. From the work by Triantafyllopoulos (2003) we know that if $N_r$ is odd, the central $N_r$-th moments are zero, and if $N_r$ is even ($N_r = 2k$) any $N_r$-th

14. The actual computational cost will be higher for the efKF filter since additional operations must be performed to link the observation and the parameter space, but these operations will similarly be cubic or lower functions of the state space dimension.

central moments can be decomposed into a sum over products of $k$ covariance terms. For example, for a four-dimensional Gaussian, one of the fourth central moments ($k = 2, 4 = 2k$) is

$$E[(s_1 - \mu_1)(s_2 - \mu_2)(s_3 - \mu_3)(s_4 - \mu_4)] = \sigma_{12}\sigma_{34} + \sigma_{14}\sigma_{23} + \sigma_{13}\sigma_{24} = \sum_{1,2,3,4} \sigma_{ij}\sigma_{kl} \quad (59)$$

where the sum is taken over all permutations of product pairs (in this case, $12/34, 14/23, 13/24$). For any $2k$-th central moment,

$$E[(s_{i_1} - E[s_{i_1}])(s_{j_1} - E[s_{j_1}]) \ldots (s_{i_k} - E[s_{i_k}])(s_{j_k} - E[s_{j_k}])] = \sum \sigma_{i_1 j_1}\sigma_{i_2 j_2} \ldots \sigma_{i_k j_k} \quad (60)$$

where the sum is again taken over all permutations of product pairs. This sum yields $(N_r - 1)!/(2^{k-1}(k-1)!)$ terms which consist of covariance elements to the power of at most $k$. For a particular central moment, this cost is independent of the dimension of the state space. Therefore the cost is dominated by the number of terms, which grows at slightly less than $\mathcal{O}(N_r!)$. There will also be an additional cost if the original polynomial was not a central moment calculation, which will involve at most $N_r$ $D$-dimensional matrix multiplications, yielding a cost of $\mathcal{O}(N_r D^2)$. In summary, the cost of computing the expected reward when the reward is a polynomial function will be

$$\mathcal{O}(L(D^3 + N_r! + N_r D^2)). \quad (61)$$

#### 5.2.3 COMPLEXITY OF CONDITIONAL MACRO-ACTION PLANNING (PBD)

Sampling beliefs from the posterior distribution over beliefs requires sampling from a multivariate Gaussian over the distribution of belief means, which we accomplish by computing the Cholesky decomposition of the covariance matrix, $\Sigma = AA^T$, an $\mathcal{O}(D^3)$ operation. Each belief mean is generated by first constructing a $D$-dimensional vector $q$, consisting of $D$ independent samples from a standard (scalar) normal distribution. A sample from the desired multivariate Gaussian $\mathcal{N}(s|\mu, \Sigma)$ is simply $\mu + Aq$. Sampling $N_s$ times involves the one-time cost of computing the Cholesky decomposition plus the matrix-vector multiplication for each sample, yielding a cost of

$$\mathcal{O}(D^3 + N_s D^2). \quad (62)$$

This procedure is performed at every branch point in the forward search tree (in other words, at all macro-action nodes except those at the tree leaves). For concreteness, consider a horizon of two macro-actions ($\tilde{H} = 2$). After expanding out each of the $|\tilde{\mathcal{A}}|$ macro-actions, we will sample $N_s$ beliefs. From each resulting belief state, we will again expand each of the $|\tilde{\mathcal{A}}|$ macro-actions: refer back to Figure 4 for an illustration. The computational complexity is now the sum of the cost at horizon one and two:

$$\mathcal{O}(|\tilde{\mathcal{A}}|(LD^3N_r + N_sD^2 + D^3) + |\tilde{\mathcal{A}}|^2 N_s LD^3 N_r) = \mathcal{O}(|\tilde{\mathcal{A}}|(N_sD^2 + D^3) + |\tilde{\mathcal{A}}|^2 N_s LD^3 C), \quad (63)$$

where the second expression is derived by considering only the higher order terms. In general, the computational complexity of selecting an action using PBD when considering a future horizon of $\tilde{H}$ macro-actions is

$$\mathcal{O}(|\tilde{\mathcal{A}}|^{\tilde{H}-1} N_s^{\tilde{H}-2}(N_sD^2 + D^3) + |\tilde{\mathcal{A}}|^{\tilde{H}} N_s^{\tilde{H}-1} LD^3 C). \quad (64)$$

| Algorithm | Computational Complexiy |
|---|---|
| PBD with Analytic Expected Reward | $\mathcal{O}(\lvert\tilde{\mathcal{A}}\rvert^{\tilde{H}-1} N_s^{\tilde{H}-2}(N_s D^2 + D^3) + \lvert\tilde{\mathcal{A}}\rvert^{\tilde{H}} N_s^{\tilde{H}-1} L D^3 C)$ (Eqn. 64) |
| PBD with Arbitrary Reward Model | $\mathcal{O}(\lvert\tilde{\mathcal{A}}\rvert^{\tilde{H}} N_s^{\tilde{H}} L D^3 + \lvert\tilde{\mathcal{A}}\rvert^{\tilde{H}} N_s^{\tilde{H}} L D^2)$ (Eqn. 66) |

Table 1: Computational complexity of selecting an action using PBD algorithm and closely related alternatives. $D$ is the number of state dimensions, $\tilde{H}$ is the macro-action forward search horizon, and $N_s$ is the number of sampled beliefs. Slightly abusing notation, we also use $N_s$ to represent the number of sampled states, in the case of arbitrary reward models.

#### 5.2.4 Complexity of PBD with Arbitrary Reward Models

For arbitrary reward models it will not be possible to analytically compute the expected reward. Instead the expected reward for each primitive action $a$ within the macro-action $\tilde{a}$ can be approximated by sampling $D$-dimensional states and estimating the expected reward by averaging the reward of each sampled state.[15] The cost of sampling $N_s$ states from a multivariate Gaussian is an $\mathcal{O}(D^3 + N_s D^2)$ operation (from Equation 62). Assuming that calculating the reward for each sample takes time linear in the state dimension, then sampling rewards adds an additional

$$\mathcal{O}(D^3 + N_s D^2 D) = \mathcal{O}(D^3(N_s + 1)) \tag{65}$$

cost to each primitive action within a macro-action, yielding a total complexity of PBD planning with reward sampling of:

$$\mathcal{O}(|\tilde{\mathcal{A}}|^{\tilde{H}} N_s^{\tilde{H}} L D^3 + |\tilde{\mathcal{A}}|^{\tilde{H}} N_s^{\tilde{H}} L D^2). \tag{66}$$

## 6. Experimental Results

In this section we test our algorithm on planning under uncertainty problems. The PBD algorithm assumes that the transition models of the problem domains can be approximated as linear Gaussians. Our results on problems inspired by two different research communities, scientific exploration from the POMDP literature (Smith & Simmons, 2005) and target monitoring from the sensor resource management domain, suggest that numerous domains do satisfy this assumption. More generally, using a linear Gaussian dynamics models is a common approximation in the controls community, and has been used to approximate even very complex dynamics such as the physiological changes involved in glucose control for diabetics (Patek, Breton, Chen, Solomon, & Kovatchev, 2007).

Despite the different origins and state space representations of the two problems that we will shortly present results for, they both involve reasoning multiple steps into the future in order to make good decisions in a very large domain. Our PBD algorithm outperforms existing approaches in both settings. We also demonstrate our algorithm in a target monitoring problem on an actual

15. Note that if the rewards are bounded, for a given $\epsilon$ and $\delta$, sampling a sufficient number of samples $N_s = f(\epsilon, \delta)$, guarantees the estimate of the expected reward of a primitive action is is $\epsilon$-close to the true expected value, with probability at least $1 - \delta$. The proof of this is a simple application of Hoeffding's inequality (1963). If $N_s$ is set such that the estimated reward of each primitive action is $\frac{\epsilon}{L}$ close to the true expected primitive action reward with probability at least $1 - \frac{\delta}{\epsilon}$, then the triangle inequality and union bound guarantee that the expected reward of the entire length-$L$ macro-action is $\epsilon$-close to the true expected reward for the macro-action with probability at least $1 - \delta$.

helicopter platform, underscoring the applicability of our algorithm to real-world domains. In all results the macro-action search horizon $\tilde{H}$ was chosen empirically given computational constraints, as is common in forward search approaches. We explicitly explore the performance changes as the search horizon is varied in Table 3. We did not use a domain-specific estimate of the future node value of search tree leaf nodes: in some domains it may be easier to specify macro-actions than a heuristic value function, and a side benefit of PBD is to be able to efficiently search to sufficient depths such that a heuristic is not required.

### 6.1 Generic Baselines

In both problems we compare the PBD algorithm to state-of-the-art approaches from the relevant research community — POMDP planners and sensor resource management algorithms for the scientific exploration and target monitoring problems respectively.

To fully examine the impact of analytically computing the posterior distribution over beliefs, we also constructed a variety of algorithms that do not currently exist in the literature. These algorithms are given access to the same hand-coded macro-actions as those used by the PBD algorithm. We first constructed comparison algorithms which use a macro-action forward search but sample observation trajectories rather than working with a posterior distribution over beliefs. Sampling observation sequences produces a particle approximation of the resulting distribution over beliefs, thereby providing a baseline algorithm that does not use an analytic representation of the posterior belief distribution. These algorithms are referred to as the macro-action discrete (MAD) algorithm when the underlying state space is discrete, and the macro-action continuous (MAC) algorithm when the state space is continuous.

We also implemented an offline point-based POMDP solver that was given access to the macro-actions used by the forward search algorithms.[16] Specifically, we modified the state-of-the-art POMDP planner SARSOP (Kurniawati et al., 2008) algorithm from the Approximate POMDP Planning (APPL) Toolkit[17] and incorporated macro-actions to guide the sampling of belief points that are used for the point-based value backups. Instead of the SARSOP algorithm using performance bounds to guide the sampling of the point-based beliefs, the modified SARSOP algorithm uses a macro-action and a sampled, same-length observation sequence to generate additional point-based belief samples. This implementation is also a modified version of the MiGS (Kurniawati et al., 2009) by the same authors. However, due to the offline, point-based nature of this modified algorithm, we were only able to evaluate the algorithm on two of the five problem domains used in this paper.

Finally, we considered an experimental comparison to an open-loop version of PBD, in which no conditioning on the received observations is ever performed; however, initial experiments suggested that this variant performed very poorly in our domains of interest, and so we did not explore it further.

### 6.2 Rocksample

The scientific exploration ROCKSAMPLE problem is a benchmark POMDP problem proposed by Smith and Simmons (2005), and subsequently extended to the FieldVisionRockSample (FVRS)

16. For a formal discussion of the differences between the offline point-based and online forward search POMDP algorithms, we refer the reader to the survey paper by Ross et al. (2008a).
17. Approximate POMDP Planning Toolkit. `http://bigbird.comp.nus.edu.sg/pmwiki/farm/appl/`

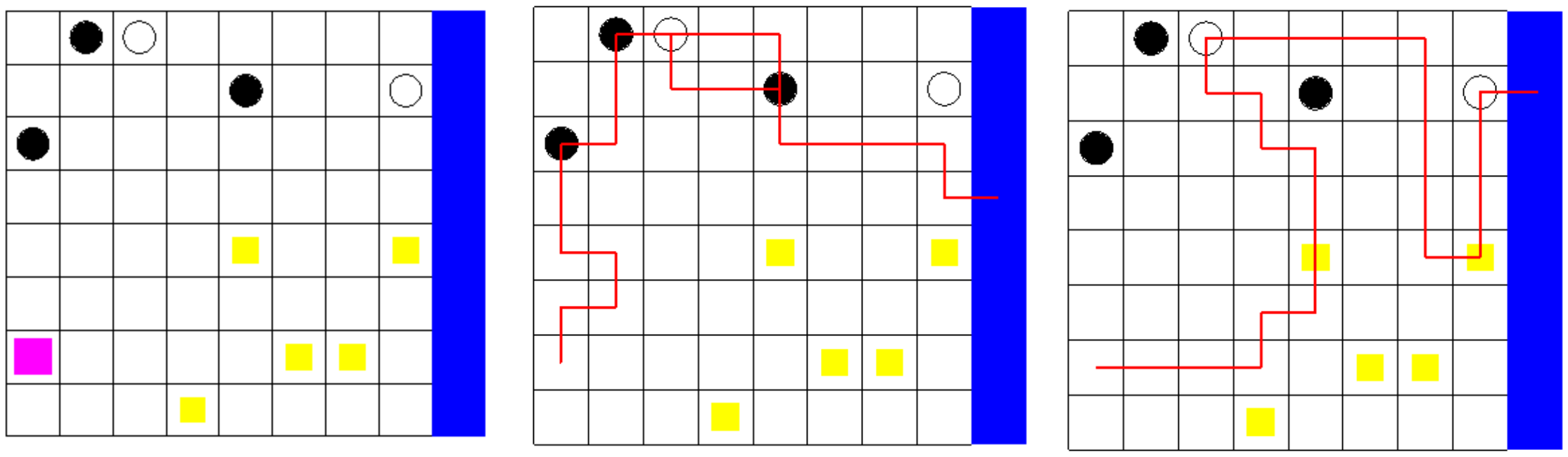

(a) ISRS(8,5) (b) SARSOP policy (c) PBD policy

Figure 5: Information Search Rocksample (ISRS) problem. (a) Initial (hidden) problem state. An agent (pink square) explores and samples rocks (circles) in the world. White circles correspond to rocks with positive value, black otherwise. Yellow squares indicate locations of the rock information beacons. The blue sidebar is the exit region. Red lines indicate paths taken by an agent executing the (b) SARSOP and (c) PBD policies. We see that the SARSOP policy only explores rocks and not the beacons; it cannot search far enough ahead to model the value of the beacons. In contrast, the PBD plan visits the beacons and then heads directly for the high-value rocks.

problem by Ross and Chaib-draa (2007). Initial experiments in these domains revealed that searching only to a shallow depth was sufficient to obtain good policies. As our interest is in domains which require long-horizon lookahead, we created a new variant of the ROCKSAMPLE problem called the Information Search Rocksample (ISRS) problem, shown in Figure 5(a). In ISRS an agent explores and samples $k$ rocks in a $n \times n$ grid world. The positions of the agent (pink square) and the rocks (circles) are fully observable, but the value of each rock (good or bad) is unknown to the agent. At every time step, the agent receives a binary observation of the value of each rock. The accuracy of this observation depends not on the agent's proximity to the rocks themselves but on the agent's proximity to rock information beacons (yellow squares), each of which correspond to a particular rock (for example, information beacons could be mountain tops that offer a particularly good view of a far off geologic formation). A key characteristic of ISRS that is not present in ROCKSAMPLE or FVRS is that the rock information beacons are not at the same locations as the rock themselves. Unlike previous ROCKSAMPLE formulations, information gathering and reward exploitation require different actions in ISRS.

The agent gets a fixed positive reward for collecting a good rock (white circle), a negative reward for collecting a bad rock (black circle), and a smaller positive reward for exiting the problem (the blue sidebar on the right). A discount factor $\gamma = 0.99$ encourages the agent to collect rewards sooner. All other actions have zero rewards.

The observation model is a Bernoulli distribution with the noise of the distribution scaled with the distance to the beacon, such that:

$$p(z_{i,t}|s_i, r_t, RB_i) = \begin{cases} 0.5 + (s_i - 0.5)2^{\frac{-\|r_t - RB_i\|_2}{D_0}} & z_{i,t} = 1 \\ 0.5 - (s_i - 0.5)2^{\frac{-\|r_t - RB_i\|_2}{D_0}} & z_{i,t} = 0 \end{cases} \tag{67}$$

where $z_{i,t}$ is a binary $\{0 \text{ or } 1\}$ observation for the value of the rock $i$ at time $t$,
$s_i$ is the true value $\{0 \text{ or } 1\}$ of the rock,
$r_t$ is the agent's position at time $t$,
$RB_i$ is the location of the information beacon associated with rock $i$,
$D_0$ is a tuning parameter that controls how quickly the accuracy of the observations decrease with greater distance between the agent and the beacon.

For example, at an information beacon, the agent, with absolute certainty, receives an observation that matches the true value of the corresponding rock, whereas when the distance between the agent and the beacon is infinite, the agent receives an "accurate" observation with 0.5 probability.

All variants of the ROCKSAMPLE problem, including our new ISRS problem, are formulated with discrete state, action and observation sets. To allow the use of our PBD and MAC algorithms, we approximate the agent's belief of each rock's value as a Gaussian distribution over the [0,1] state space, and take advantage of the efKF presented in Section 4 to represent the ROCKSAMPLE problem's Bernoulli observation model (Equation 67: see Appendix B for details).

Each macro-action is a finite, open-loop sequence of primitive actions. For the ROCKSAMPLE problem, there are five primitive actions: single steps in the four cardinal directions and the rock sampling action. Recall that the agent's position is fully observable and its actions are deterministic. Using domain knowledge, the macro-actions considered from a particular belief state are macro-actions that, given the agent's current position, consist of a sequence of actions that enables the agent to move to each rock, each information beacon, or to the nearest exit. This results in $2k + 1$ macro-actions being considered for forward search at every belief node. As the agent operates in a grid world, there may be multiple action sequences with the same, shortest distance between two grid squares: the macro-action considered is the one where the agent would move as diagonally as possible, so as to replicate the agent's shortest path movement in a continuous map. In addition, if the agent is currently on a rock (which is fully observable), additional macro-actions where the agent first collects the rock before executing one of the $2k+1$ default macro-actions are considered, resulting in twice as many macro-actions. The set of macro-actions therefore varies with every belief node.[18] For an ISRS problem with 5 rocks in a $8 \times 8$ grid world, the average macro-action length was $4.76$, with a minimum and maximum macro-action length of 1 and 12 respectively.

As the ROCKSAMPLE family of problems originates from the POMDP literature, we compared our macro-action algorithms to existing state-of-the-art POMDP solvers: the fast upper-bound of QMDP (Littman, Cassandra, & Kaelbling, 1995), the point-based offline value-iteration techniques HSVI2 (Smith & Simmons, 2005) and SARSOP (Kurniawati et al., 2008), as well as RTBSS (Paquet, Chaib-draa, & Ross, 2006), an online, factored, forward search algorithm. We also evaluated a modified version of the SARSOP algorithm that was given access to the same macro-actions used by the forward search macro-action algorithms. Since all approaches, including our own, are approximations, we also include as an upper bound the value of the fully observable problem.

Table 2 compares the performance of the different algorithms in the ISRS problem. Each algorithm was tested on 10 different initial conditions (which rocks were high valued and which were low valued), and each scenario was tested 20 times. The HSVI2 and SARSOP algorithms were executed offline for a range of durations,[19] while the forward search algorithms were allowed to search

18. However, if two belief nodes have the same agent position, their macro-actions will be identical.
19. The offline execution durations for both HSVI2 and SARSOP were chosen empirically. HSVI2 was able to search for solutions to the ISRS[8,5] problem for 1,000s offline before running out of memory. It was found that the values computed by SARSOP remained constant after 25,000s.

| | ISRS[8,5] | | |
|---|---|---|---|
| | Avg rewards | Online time (s) | Offline time (s) |
| QMDP | $1.11 \pm 0.15$ | 0.0001 | 3.03 |
| HSVI2 | $6.78 \pm 0.62$ | 0.051 | 1000 |
| SARSOP | $8.46 \pm 0.70$ | 0.070 | 25000 |
| SARSOP(macros) | $18.78 \pm 1.59$ | 0.015 | 1000 |
| RTBSS (d5, s10) | $9.78 \pm 0.49$ | 17.64 | 0 |
| RTBSS (d7, s2) | $12.41 \pm 0.46$ | 3.28 | 0 |
| RTBSS (d10, s1) | $15.39 \pm 0.45$ | 7.0357 | 0 |
| MAC (d3,s50) | $13.68 \pm 0.65$ | 15.39 | 0 |
| MAD (d3,s50) | $15.88 \pm 0.54$ | 4.81 | 0 |
| PBD (d3,s50) | $14.76 \pm 0.57$ | 1.26 | 0 |
| Fully observable | 21.37 | N.A. | N.A. |

Table 2: ISRS results. HSVI2 and SARSOP were executed offline for a range of durations. For the forward search algorithms, the numbers in brackets represent the search depth (d) and number of posterior beliefs obtained (s) at the end of each action/macro-action. Online time indicates the average time taken by the planner to return a decision at every time step. Standard error values are shown.

out to pre-defined depths. Here, depth refers to the primitive action depth in the RTBSS algorithm, and the macro-action depth in the macro-action algorithms (MAC, PBD and MAD). In addition, a pre-defined number of samples were used to obtain posterior beliefs after every action/macro-action. We abuse notation here slightly by using samples to refer to observations in the RTBSS algorithm, observation sequences in the MAD and MAC algorithms, and to samples from the posterior belief distribution in the PBD algorithm.

We also attempted to allow the RTBSS algorithm to search to the same primitive action search depth as the macro-action algorithms do on average, i.e. $4.76 \times 3 \approx 14$, by reducing the number of observations that are sampled per action. We found that even if only 1 observation was sampled per action, RTBSS could only achieve a search depth of 10 in reasonable computation time.

The macro-action algorithms do significantly better than most of the other benchmark solvers. Figure 5(b) and 5(c) compare the policies generated by the SARSOP algorithm and the PBD algorithm in the ISRS problem. Both SARSOP and HSVI2 explore parts of the belief space guided by an upper bound on belief-action values. A long lookahead is required to realize that visiting beacons and then rocks has a higher value that visiting rocks, so many iterations and therefore substantial computation time is required for SARSOP and HSVI2 to sample the beliefs that will lead to them computing a higher-value policy. In the considerable offline computation time provided, both SARSOP and HSVI2 did not discover that it is valuable for the agent to make a detour to the information beacons before approaching the rocks. Instead, they directly approach the rocks and make decisions based on the noisy observations that are obtained due to the large distance from the information beacons.

The RTBSS algorithm does reasonably well when it is able to search deep enough, once again emphasizing the need for planning under uncertainty algorithms to search far into the future in order to perform well. Nevertheless, when the same amount of online planning time is available, the MAD algorithm still outperforms the RTBSS algorithm. Macro-actions allow the algorithms to uncover the potential value of moving to an information beacon without incurring the computational cost of primitive-action forward search; this allows our macro-action forward search approaches to perform better than prior primitive-action approaches. Figure 5(c) shows that a PBD agent's policy involves visiting some of the information beacons to gather information about which of the rocks are good (white circles), before traveling to those rocks to sample them. In this domain, MAD does better than the PBD algorithm since the problem specification is made up of discrete states, whereas the parametric approaches must approximate the world models during planning. In addition, the fully-factored nature of the problem domain, where the state of each rock value is independent, keeps the computational requirements of the MAD algorithm relatively small.

| | Depth 1, Samples 50 | | Depth 2, Samples 50 | | Depth 3, Samples 50 | | Depth 4, Samples 20 | |
|---|---|---|---|---|---|---|---|---|
| | Avg rewards | Online time(s) | Avg rewards | Online time(s) | Avg rewards | Online time(s) | Avg rewards | Online time(s) |
| MAC | 4.61±0 | 0 | 9.63±0.77 | 0.022 | 13.68±0.65 | 15.39 | 15.07±1.62 | 660.50 |
| MAD | 4.61±0 | 0 | 7.51±0.84 | 0.0083 | 15.88±0.54 | 4.81 | 17.43±0.78 | 225.74 |
| PBD | 4.61±0 | 0 | 7.73±0.77 | 0.002 | 14.76±0.57 | 1.26 | 15.82±0.77 | 75.06 |

Table 3: Performance of macro-action algorithms with different macro-action depth on ISRS. At depth 4, a smaller number of posterior beliefs were sampled for computational reasons.

Similarly, when the SARSOP algorithm was modified to incorporate the hand-coded macro-actions, this offline, point-based algorithm performed much better than existing offline approaches, including the SARSOP algorithm without access to macro-actions. This result re-emphasizes that well-designed macro-actions can be very valuable in generating good policies in partially observable domains. However, not all problem domains, especially those with large, factored domains that are of interest in this paper, can be represented and solved in an offline manner, and we shall shortly see the benefit of PBD for such settings.

We also performed additional analysis on the three macro-action forward search algorithms. Table 3 compares the different rewards obtained by the macro-action algorithms for different macro-action depths, as well as the time taken by the planner to return a decision at every time step. The sharp performance jump that occurs when the macro-action search depth is increased from 2 to 3 emphasizes the need to search to a longer horizon in the ISRS problem before a good policy can be generated. However, the computational cost of the algorithms also increases exponentially with the macro-action search depth. This table also illustrates the small loss in performance induced by approximating the discrete problem with the continuous representation of either MAC or PBD, and the substantial increase in computational speed using PBD.

Next we examine the relative performance and computational cost of PBD, MAC and MAD, as the number of samples changes (Table 4) up to a search depth of 3. Recall that the PBD algorithm samples from the posterior belief at each node in the search tree, and evaluates the expected future reward of subsequent macro-actions for each sample. Different regions of the posterior belief space may plan to use different subsequent macro-actions, allowing the planner to implicitly condition its plans on the received observations. However, the sampling used to partition the posterior belief space and assign different actions to different beliefs introduces a source of approximation error and additional computational complexity. As predicted by our earlier computational complexity analysis, PBD scales best of the three algorithms as the number of samples increases, since it does

| | 5 Samples | | 50 Samples | | 100 Samples | | 500 Samples | |
|---|---|---|---|---|---|---|---|---|
| | Avg rewards | Online time(s) | Avg rewards | Online time(s) | Avg rewards | Online time(s) | Avg rewards | Online time(s) |
| MAC | 12.76±0.54 | 0.15 | 13.68±0.65 | 15.39 | 12.47±0.70 | 58.90 | 12.94±2.57 | 1732.52 |
| MAD | 15.31±0.52 | 0.056 | 15.88±0.54 | 4.81 | 15.57±0.66 | 20.72 | 16.32±2.18 | 552.64 |
| PBD | 12.92±0.57 | 0.035 | 14.76±0.57 | 1.26 | 14.56±0.59 | 4.52 | 15.36±1.15 | 108.64 |

Table 4: Performance of macro-action algorithms in ISRS up to depth 3 with different numbers of samples.

(a) ISRS(15,6)

| | Avg rewards | Online time (s) | Offline time (s) |
|---|---|---|---|
| SARSOP | 9.43 ± 1.03 | 0.00006 | 10000 |
| SARSOP(macros) | 11.42 ± 0.49 | 0.00006 | 900 |
| RTBSS(d7,s2) | 8.37 ± 0.55 | 4.98 | 0 |
| RTBSS(d10,s1) | 9.35 ± 0.65 | 10.91 | 0 |
| MAC(d3,s20) | 15.94 ± 0.92 | 7.01 | 0 |
| MAD(d3,s20) | 17.57 ± 0.82 | 2.74 | 0 |
| PBD(d3,s20) | 17.00 ± 0.83 | 0.58 | 0 |
| Fully obs. | 30.95 | N.A. | N.A. |

(b) ISRS(100,30)

| | Avg rewards | Online time (s) |
|---|---|---|
| SARSOP | N.A. | N.A. |
| SARSOP(macros) | N.A. | N.A. |
| MAC(d3,s5) | 42.64 ± 3.78 | 310.05 |
| MAD(d3,s5) | 51.70 ± 3.46 | 101.92 |
| PBD(d3,s5) | 43.68 ± 2.00 | 60.81 |
| Fully obs. | 66.61 | N.A. |

Table 5: Performance on larger ISRS problems

not have to perform belief updates along each sampled trajectory explicitly. In general, performance improves with more samples, although the improvement was not statistically significant in the ISRS problem. However, when a decision-making under uncertainty problem requires a large number of posterior beliefs to be sampled after every macro-action, the PBD algorithm results in consistently faster performance for the same number of samples. Once again, MAD has a slight performance edge due to the approximation of the discrete ISRS problem with continuous variables implicit in PBD, but the difference is again not significant.

The macro-action forward search nature of our algorithm also allows us to scale to much larger versions of the ROCKSAMPLE problem, since unlike offline techniques, it is unnecessary to generate a policy that spans the entire belief space. We compared the algorithms on two additional ISRS problems — a 16 by 16 grid with 6 rocks, and a 100 by 100 grid with 30 rocks.

Both problem domains were too large for most of the benchmark solvers that were originally used for comparison, though the SARSOP and RTBSS algorithms could be implemented for the ISRS[15,6] problem domain. Table 5(a) shows the performance of SARSOP and the forward search algorithms for the ISRS[15,6] problem domain. The modified SARSOP algorithm that incorporates macro-actions ran out of memory after computing a policy offline for 900s. Because the forward search macro-action algorithms are better able to concentrate computational resources on the reachable belief space from the agent's current belief, the forward search macro-action algorithms perform much better than both the SARSOP algorithm and the modified version that incorporates macro-actions. Similarly, while the forward search single-action RTBSS algorithm performed reasonably well on the ISRS[8,5] problem if the search depth was sufficiently large, the algorithm was unable to search sufficiently deep in reasonable time on the larger ISRS[15,6] problem, resulting in poorer performance than the forward search macro-action algorithms.

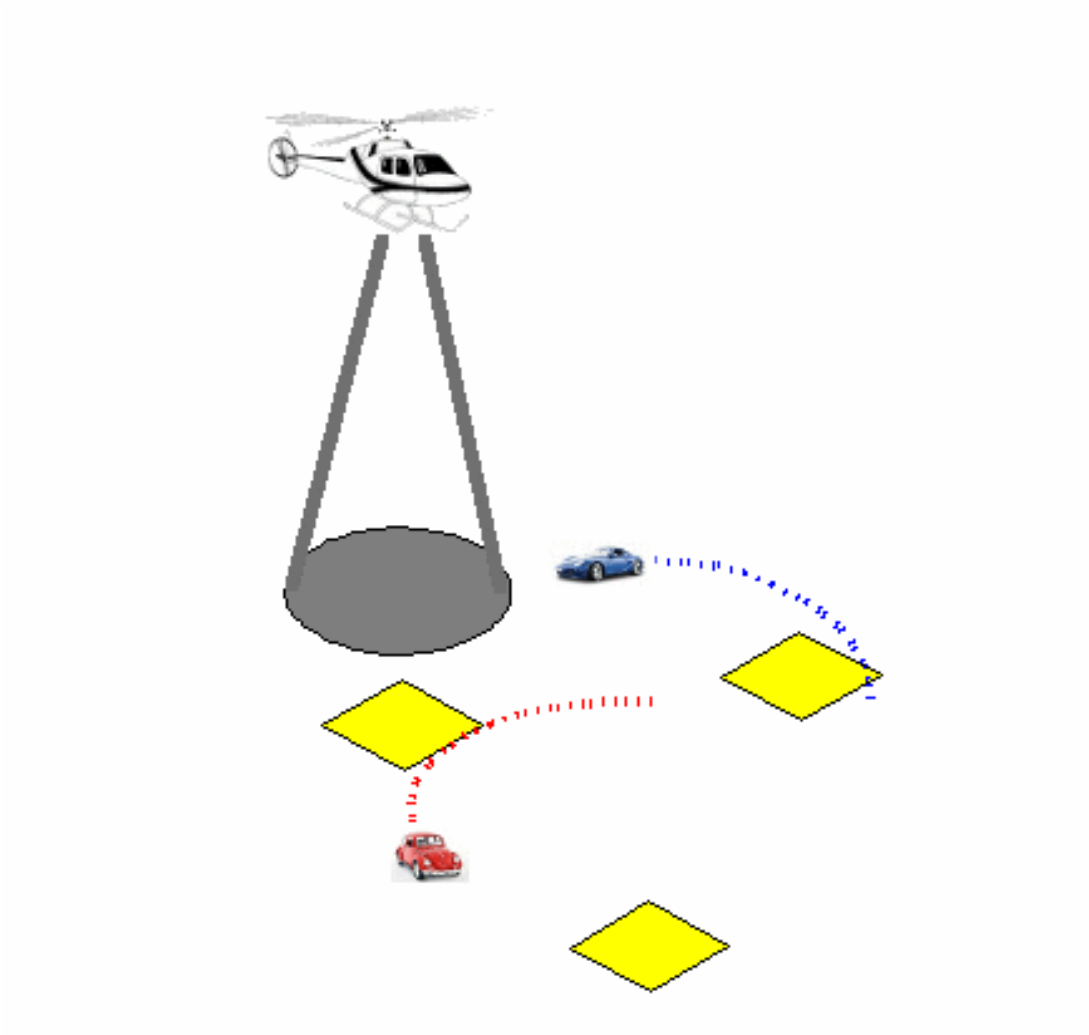

Figure 6: TARGETMONITOR problem. A helicopter must track multiple targets moving with noisy dynamics. The field-of-view of the agent's sensor (shaded circle) increases with the agent's altitude.

We further implemented the macro-action algorithms on a ISRS[100,30] problem domain, which far exceeds any problem that can be solved by a traditional POMDP solver, including the modified SARSOP algorithm that incorporates macro-actions. Table 5(b) compares the results of the three macro-action algorithms to the fully observable value, which provides a strict upper bound of the maximum possible reward for the problem. Such large problems also underscore the value of having macro-actions to limit the branching factor of the forward search.

### 6.3 Target Monitoring

We next consider a target monitoring problem related to those studied in the sensor resource management literature (Scott, Harris, & Chong, 2009). In this problem (Figure 6), a helicopter agent has to track multiple targets that are moving independently with noisy dynamics. The helicopter operates in 3D space, while the targets move on the 2D ground plane. The helicopter is equipped with a downward-facing camera for monitoring the environment, and if a target is within the field-of-view of the camera sensor, the agent receives a noisy observation of the location and orientation of the target. We assume for simplicity that the observations of each target are unique, allowing us to ignore the data association problem that has been addressed elsewhere.

The noise associated with the agent's observation of a target depends on the agent's position relative to the target. When the helicopter is close to the ground it can only observe a small region, but can determine the position of objects within that small region to a high level of accuracy. When the helicopter flies at a higher altitude, it can view a wider region of the environment, but its measurements will be less precise. Similarly, the closer the helicopter is to a particular target, the more accurate the helicopter's observation of that target is expected to be. Reflecting this intuition, we use a Gaussian observation model where the noise covariance is a function of the position of the helicopter and target: details of this sensor model are provided in Appendix C. One desirable attribute of our sensor model is that if the helicopter is very uncertain about a target's location, even if the helicopter is close to the target's mean location, a single observation is unlikely to localize the target. If the target location is very uncertain, there is a low probability that the target is within the helicopter's field of view.

| | 1 Target | | 2 Targets | |
|---|---|---|---|---|
| | Avg rewards | Online time (s) | Avg rewards | Online time (s) |
| Greedy | -21.50 ± 7.65 | 0.0657 | -26.50 ± 5.00 | 0.19 |
| WT-Single | 65.14 ± 8.64 | 0.00075 | -27.03 ± 8.06 | 0.00068 |
| WT-Macro | 64.64 ± 8.28 | 0.00076 | -19.05 ± 7.64 | 0.00042 |
| NBO | -5.80 ± 7.92 | 0.051 | -10.78 ± 6.95 | 0.21 |
| MAC(d2,s10) | 41.73 ± 6.96 | 4.73 | 46.67 ± 18.91 | 22.13 |
| MAD(d2,s3) | 1.27 ± 5.23 | 1.66 | 0.97 ± 5.82 | 8.46 |
| PBD(d2,s10) | 36.21 ± 6.52 | 0.89 | 68.00 ± 16.65 | 4.33 |
| | 3 Targets | | 8 Targets | |
| Greedy | -18.00 ± 7.15 | 0.46 | -95.00 ± 23.37 | 2.01 |
| WT-Single | -23.52 ± 10.89 | 0.00080 | -71.17 ± 14.53 | 0.00063 |
| WT-Macro | -10.53 ± 17.12 | 0.00037 | -52.98 ± 21.74 | 0.00025 |
| NBO | -8.27 ± 8.84 | 0.63 | -5.98 ± 18.00 | 5.78 |
| MAC(d2,s10) | 37.89 ± 12.49 | 70.91 | 83.86 ± 25.65 | 711.67 |
| MAD(d2,s3) | -1.86 ± 5.19 | 26.96 | 27.36 ± 14.74 | 432.13 |
| PBD(d2,s10) | 55.78 ± 13.84 | 13.02 | 120.80 ± 25.77 | 132.50 |

Table 6: TARGETMONITOR Results. Run for 200 time steps.

The agent's pose is fully observable, though the actions that it takes are subject to a small amount of additive Gaussian noise. As a result, unlike the ROCKSAMPLE domains, the open-loop nature of macro-actions means that the planner cannot perfectly predict the vehicles' pose at the end of the macro-action. Each target's motion is determined by its translational and rotational velocities. The model provides the agent with a prior over these velocities, but at every time step, the target's true velocities are additive functions of these fixed input controls and Gaussian noise. In the parametric formulation, the agent maintains a Gaussian belief over each target's state, and in order to compare MAD, we discretize the continuous state spaces of the agent's and targets' positions, and maintain a probability distribution over each discrete target state. Due to computational memory constraints, for a $100m$ by $100m$ by $20m$ target monitoring problem in the $x, y$ and $z$ directions, we were limited to a discretization with $10m$ resolution in the $x, y$ directions, $5m$ in the $z$ direction, and $45°$ angular resolution.

We focus on a particular decision-theoretic version of the sensor resource management problem, where at each time step the agent must decide if each of the targets is inside an area of interest. These areas of interest are indicated by the yellow squares in Figure 6. The agent receives a positive reward if it successfully reports that a target is in an interest region, a negative reward if it wrongly decides that the target is in the region, and no reward if it decides that the target is not in the region, regardless of the target's actual state. Small costs are incurred for the agent's motion. We call this the TARGETMONITOR problem.

Given the current location of the agent, macro-actions were generated by computing the sequence of actions that will enable the agent to move to a particular altitude over the means of each target belief. For a particular desired destination, a macro-action is constructed by first computing the shortest path between the agent's current and desired location, and then dividing this path into primitive actions based on the maximum length of each primitive action. We also included a hovering macro-action that consists of hovering at the agent's current location for four time steps. Note that the agent's current location is fully-observable, and for the purpose of generating macro-actions, we assume that the primitive actions are noise-free. Hence, for each primitive action, the helicopter is assumed to move by the mean expected change. Similar to the ROCKSAMPLE problem, although the macro-actions are generated according to a policy that relies on domain knowledge, the macro-actions themselves are evaluated in the forward-search algorithms as open-loop sequences of primitive actions. We compare the forward search macro-action algorithms to a range of intuitive strategies and prior approaches. The first algorithm is the greedy strategy, which returns the primitive action that results in the largest expected reward in the next step. The next two approaches are the Worst Target (WT) policies, which are hand-coded policies of traveling to the target that has the largest uncertainty of all the targets being tracked. The intuition is that the agent's goal in general is to localize the targets in the environment. The two algorithms differ based on whether the agent chooses a new target to travel to after each time step (WT-single), or re-plans only after it has reached the target it had initially chosen (WT-macro). Finally, we compared our algorithm to the nominal belief optimization (NBO) algorithm proposed by Scott, Harris and Chong (2009). The NBO algorithm also assumes a Kalman filter model for the target monitoring problem, but rather than considering the entire distribution of posterior beliefs, only the most likely posterior belief after an action is considered. In this algorithm, the most likely posterior belief for a Gaussian belief update is given by the posterior mean without incorporating any observations, and the covariance given by linearizing about the most likely mean at each step. Although the original algorithm uses an optimization approach to search for action sequences, here we modify the NBO algorithm by adopting a forward search approach, evaluating each macro-action based on the most likely posterior belief.[20]

Table 6 presents results for the TARGETMONITOR problem, comparing the algorithms in scenarios with different number of targets. These results demonstrate that the PBD algorithm, with its closed form representation of the distribution of posterior beliefs after an action, finds a significantly better policy than alternate approaches. Figure 7 demonstrates a typical policy executed by the PBD algorithm. The agent begins in the middle of the grid world, and approaches a target at a high altitude (Figure 7(b)), maximizing the likelihood of localizing that target. If none of the targets seem to be approaching a region of interest, the agent hovers in the same position to conserve energy (Figure 7(c)). When one of the targets may potentially be entering a region of interest, the agent focuses on that target, tracking it carefully to ensure that it knows when the target is exactly in the region of interest (Figure 7(d),(e)). The agent subsequently travels to a high altitude and repeats the process of localizing another target with potential rewards (Figure 7(f)).

Considering the entire distribution of posterior beliefs, rather than just the maximum likelihood posterior belief, is valuable because the agent is able to reason that there is a possibility that the

20. As noted by the authors, the NBO algorithm focuses on a new method for approximating the Q-value, rather than on the optimization techniques. While they adopt a generic search approach for performing the optimization, the authors also point to forward-search POMDP algorithms as good search techniques in which their Q-value approximations could be incorporated. Our use of forward search with the NBO Q-value approximation does not affect the results.

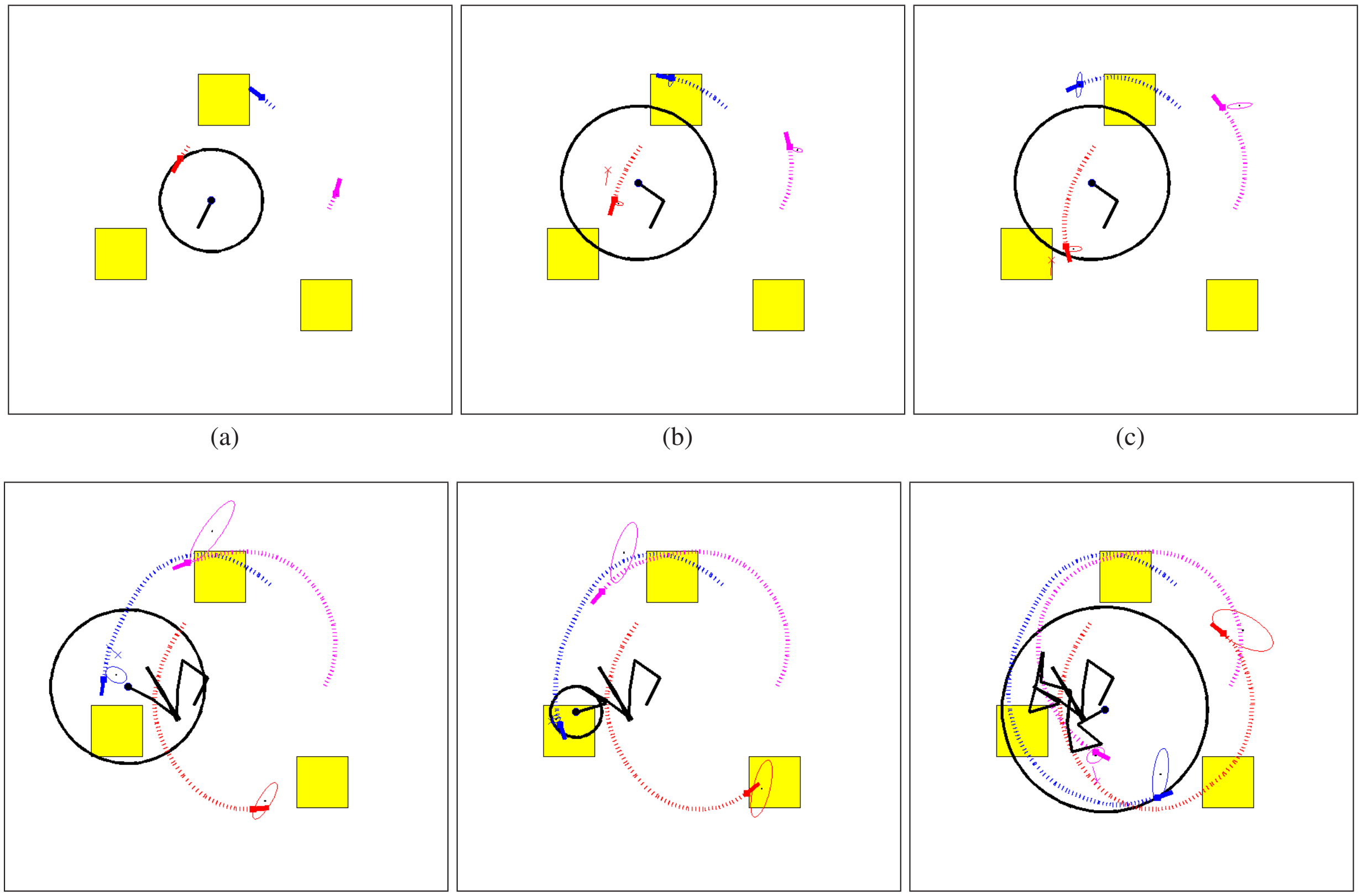

Figure 7: Snapshots of the PBD policy being executed. The black circle indicates the field-of-view of the agent's sensor, which is directly proportionate to the agent's height. The size of the error ellipses indicate the agent's uncertainty associated with each target at each time step. The agent alternates between flying at a high altitude to maximize the likelihood of observing targets (b),(f) and focusing on a single target that is near/has entered an area of interest (e).

target could be within a region of interest. In contrast, the NBO approach only considers the most likely posterior belief, and will seek to localize the target only if the mean of its belief appears to be heading into a region of interest. While the consideration of the entire distribution of posterior beliefs necessarily incurs greater computational cost, we demonstrate in Section 6.4 that we are able to track two targets in real-time using an implementation of the PBD algorithm that has not been optimized for speed.

Table 6 also shows that because the PBD algorithm directly computes the distribution of posterior beliefs after a macro-action, the computational cost of the PBD algorithm is significantly lower than the MAC algorithm. The MAC algorithm suffers a greater computational cost as it generates the set of posterior beliefs after a macro-action by sampling observation sequences and explicitly performing belief updating along each sample trajectory. In addition, because the TARGETMONITOR problem has a state space that is fundamentally continuous, the resolution of the state space

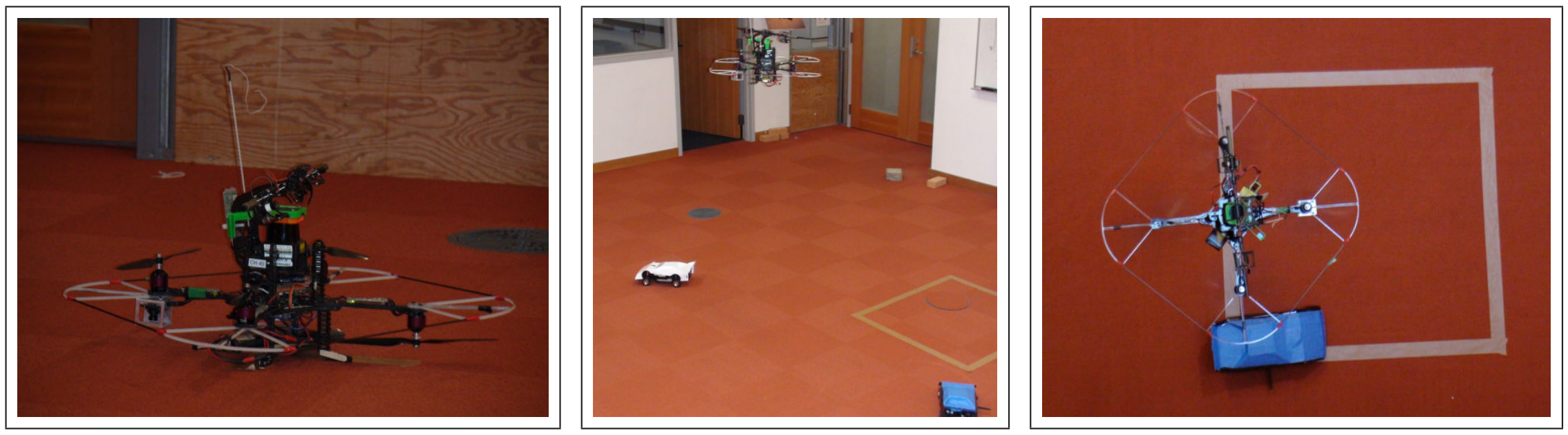

(a) Our Quadrotor helicopter (b) Multiple cars being tracked. (c) Helicopter tracking car through an area of interest

Figure 8: TARGETMONITOR demonstration with helicopter. The helicopter has to simultaneously track two cars and report whenever either car enters an area of interest.

discretization that was achievable given computational memory constraints was still unable to capture the inherent characteristics of the target monitoring problem, resulting in the poor performance of MAD in the TARGETMONITOR problem.

In the single-target case, we also observed the result that the PBD algorithm does worse than the hand-coded policy of the agent traveling to the target with the largest uncertainty (WT-single). When the problem only involves a single target, such a policy equates to having the agent hover over the sole target at every step, which is the optimal policy in the single target case. In contrast, we observe that the MAC and PBD algorithms return policies that result in the agent periodically leaving the target to fly to a higher altitude, resulting in greater noise in the observations and corresponding loss of rewards on average. By restricting the MAC and PBD algorithms to planning with macro-actions, we restrict the set of plans the agent can consider in order to search deeper, rather than a shorter conditional plan that is conditioned on the observations after each *primitive* action. Even though the agent re-plans after every time step, without this conditional plan, an agent executing the MAC or PBD algorithms will execute the “safe” policy and fly to a higher altitude, which maximizes the likelihood of keeping the target well-localized when it is unable to condition its actions based on subsequent observations. This example highlights the trade-off we make by considering a smaller class of policies (those that can be expressed as chains of macro-actions) compared to the full policy set. While in simple problems, such as a single-target TARGETMONITOR problem, the policy restriction can clearly be a limitation, our macro-action algorithms perform significantly better than the other benchmark approaches when there are multiple targets, in scenarios that are arguably more complicated and require more sophisticated planning algorithms.

### 6.4 Real-world Helicopter Experiments

Finally, as a proof of concept, we demonstrate the PBD algorithm on a live instantiation of the TARGETMONITOR problem. A motivating application for this monitoring problem is our involvement (He et al., 2010a) in the 1st US-Asian Demonstration and Assessment of Micro Aerial Vehicle (MAV) and Unmanned Ground Vehicle (UGV) Technology (MAV’08 competition). The mission was a hostage rescue scenario, where an aerial vehicle had to guide ground units to a hostage building while avoiding an enemy guard vehicle. Our aerial vehicle therefore had to plan paths in order to be able to monitor the different ground objects and report whenever any of them arrived at an area of interest.

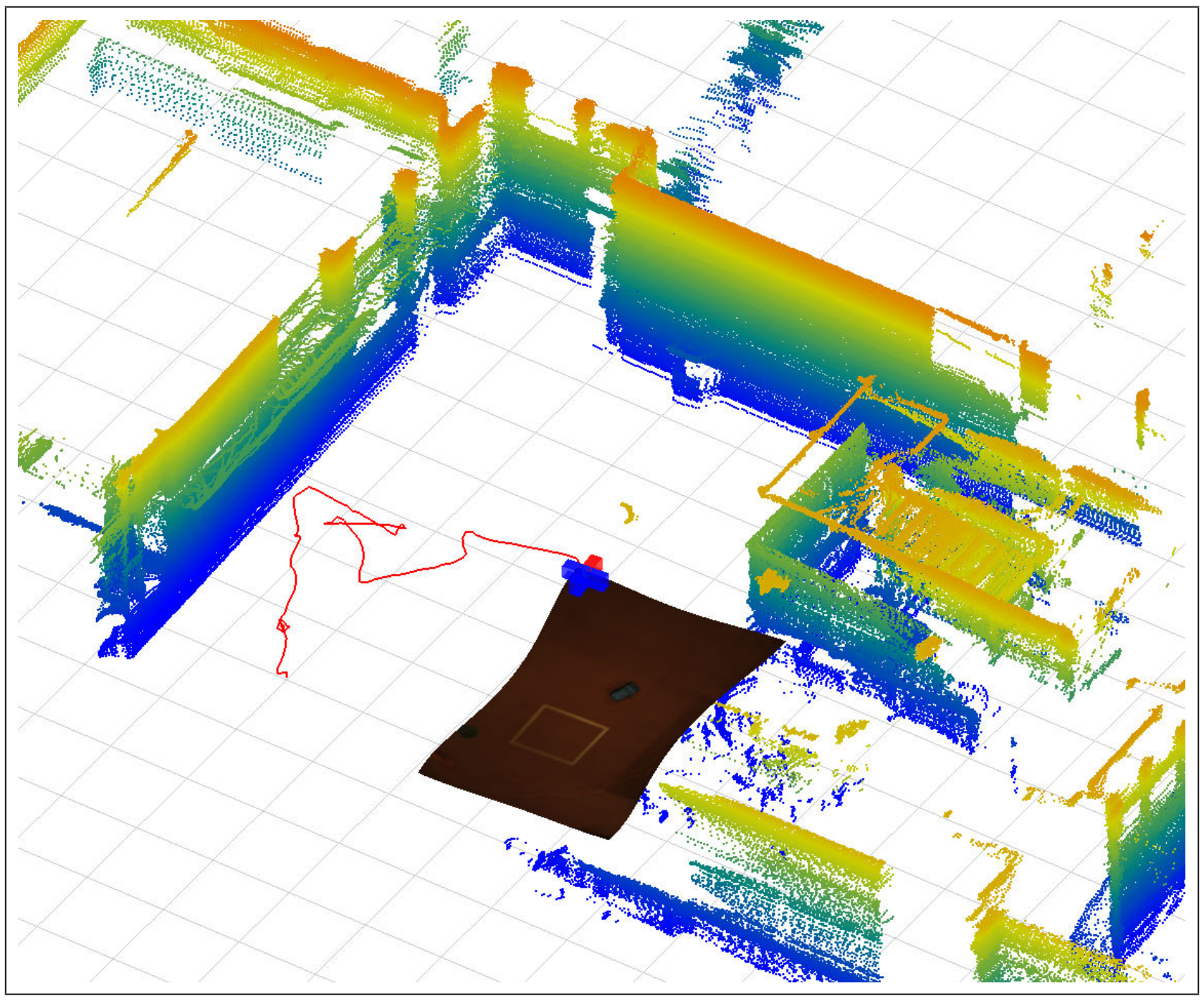

Figure 9: The helicopter (blue/red cross) uses an onboard laser scanner to localize itself. A downward pointing camera is used to observe the ground targets. In this figure, the camera image from the onboard camera is projected onto the ground plane.

We demonstrate this scenario on an actual helicopter platform monitoring multiple ground vehicles in an indoor environment (Figure 8b). In previous work (He, Prentice, & Roy, 2008; Bachrach, He, & Roy, 2009), we developed a quadrotor helicopter (Figure 8a) that is capable of autonomous flight in unstructured and unknown indoor environments. The helicopter uses a laser rangefinder to localize itself in the environment.

We mounted a downward-facing camera to make observations of the target. Since target detection is not the focus of this paper, each of the ground vehicles had a known, distinctive color, to be detected and distinguished easily with a simple blob detection algorithm. Given the helicopter's position in the world and the image coordinates of the detected object, we were able to recover an estimate of the position and orientation of a target observation in global coordinates. The helicopter only received an observation of the target when the target was within the camera's field-of-view, and although the helicopter platform hovered relatively stably, slight oscillations persisted, which resulted in noisier observations when the helicopter was flying at higher altitudes. Hence, the helicopter had to choose actions that balanced between obtaining more accurate observations at low altitudes and a larger field-of-view by flying high.

Two ground vehicles were driven autonomously in the environment with open-loop control, and the helicopter had to plan actions that would accurately localize both targets. To replicate the TARGETMONITOR problem, we marked out three areas of interest where the helicopter had to predict at every time step if the targets were within those areas (Figure 8c). We applied the PBD algorithm to plan paths for the helicopter that maximized the likelihood that it could accurately report whenever a target is in an area of interest. However, rather than sending open-loop control actions to the helicopter, as we did in the simulation experiments, for safety reasons we closed the loop around the position of the helicopter, sending desired waypoints that we wanted the helicopter to navigate to. The helicopter's true state in the world was actually partially observable, and the helicopter had to rely on an onboard laser scanner to localize its position in the environment.

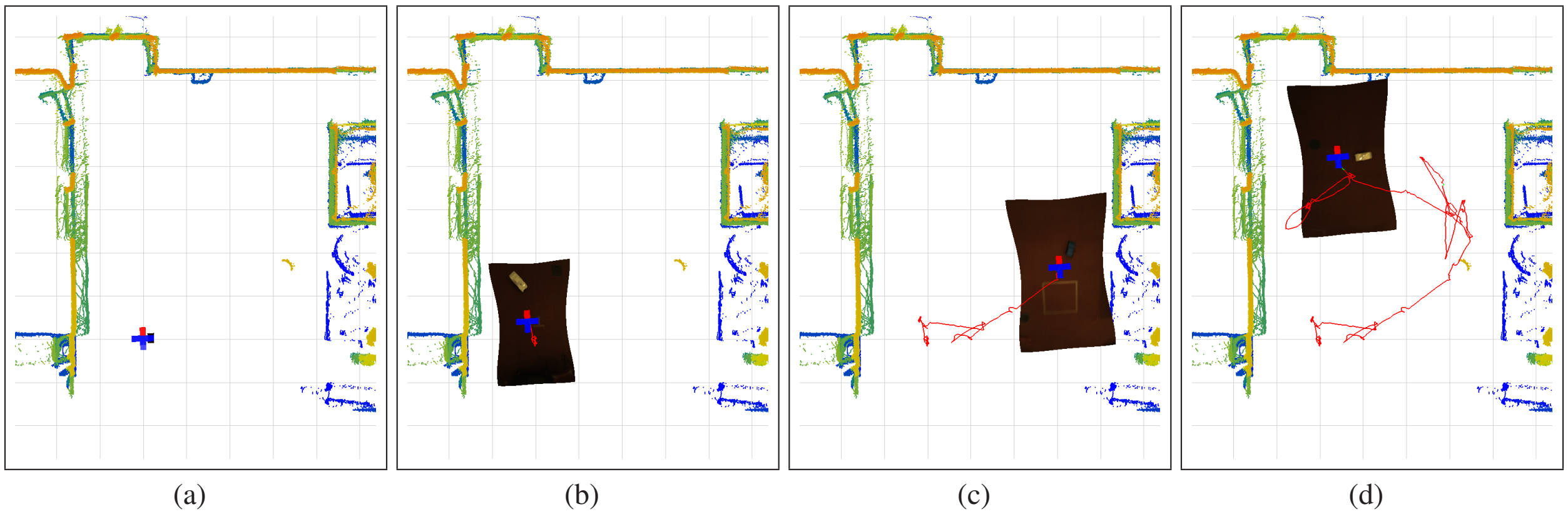

Figure 10: Bird's eye-view snapshots of the helicopter's trajectory (red), based on policy generated by the PBD algorithm. The helicopter (blue/red cross) alternates between observing the white (b,d) and blue (c) cars in order to accurately report when either car is in an area of interest. The area of the field-of-view of the agent's camera sensor varies directly with the height that the agent is flying at.

| | # Target entry detections | # True target entries | Flight time (s) | Dist. traveled (m) |
|---|---|---|---|---|
| WT-Single | 1 | 7 | 484.15 | 243.36 |
| NBO | 1 | 4 | 435.25 | 247.01 |
| PBD | 4 | 6 | 474.64 | 282.51 |

Table 7: Performance of algorithms on real-world helicopter experiment. Ground truth was found using an overhead video camera.

Figure 9 shows a 3D view of the helicopter as it monitors and reports on the locations of the ground targets. As the helicopter flew around the environment, it obtained observations of the target, which were then used to update the agent's belief of the targets. Figure 10 provides snapshots of the helicopter executing a plan that is computed online by the PBD algorithm. The helicopter exhibited similar behaviors to those that were observed in the simulation experiments. The helicopter alternated between the two targets in the environment to report when either target was in an area of interest. When the agent had a large uncertainty over a particular target's location, it would also fly to a higher altitude in order to increase its sensor field-of-view, thereby maximizing the likelihood that it will be able to re-localize the targets. A video of the complete system in action is available at: `http://groups.csail.mit.edu/rrg/index.php?n=Main.Videos`.

As a coarse measure of achieved reward, we evaluated how well the helicopter running PBD did at monitoring when a target entered an area of interest, and compared it to the WT-Single and NBO algorithms. The ground truth of the number of times the targets actually entered the areas of interests in each trial was found by using a video camera mounted overhead above the environment. Table 7 indicates that the PBD algorithm did a much better job of monitoring the targets' positions than both the WT-Single and NBO algorithms. In particular, we observed that both the WT-Single and NBO algorithms seldom took advantage of the ability to increase the agent's sensor field-of-view by having the agent fly to a higher altitude. An agent applying these two algorithms therefore had a higher probability of losing track of the targets completely.

## 7. Related Work

Decision-making under uncertainty when the states are partially observable is most commonly discussed under the Partially Observable Markov Decision Process (POMDP) framework, though this problem has also been analyzed in other research domains under similar assumptions. While it is beyond the scope of this paper to provide a comprehensive survey of POMDP techniques, point-based methods such as HSVI2 (Smith & Simmons, 2005) and SARSOP (Kurniawati et al., 2008) are often considered state-of-the-art offline methods, leveraging the piece-wise and convex aspects of the value function to perform value updates at selected beliefs. These approaches assume a discrete-state representation, but offline approaches that use parametric representations have been proposed for continuous-valued state spaces (Brooks, Makarenko, Williams, & Durrant-Whyte, 2006; Brunskill, Kaelbling, Lozano-Perez, & Roy, 2008; Porta et al., 2006). Hoey and Poupart (2005) have also addressed continuous observation spaces by finding lossless partitions of the observation space. Recent work by Bonet and Geffner (2009) suggests that alternate point-based approaches that use tabular representations of the value function may also be competitive with prior point-based approaches which used $\alpha$-vector representations, and this alternate representation may be useful for continuous domains. The ideas in this paper are more closely related to the body of online, forward search POMDP techniques that only compute an action for the current belief, which were recently surveyed by Ross et al. (2008a).

Macro-actions have been considered in depth within the fully observable Markov decision process community, and are typically known as "options" (Sutton et al., 1999), or posed as part of a semi-Markov decision process (Mahadevan, Marchalleck, Das, & Gosavi, 1997). These prior formalisms for temporally-extended actions include closed-loop policies that persist until a termination state is achieved. It would be interesting to explore in the future how these richer notions of macro-actions could be incorporated into our approach.

Several offline POMDP approaches use macro-actions such as those of Pineau, Gordon, and Thrun (2003b), Hansen and Zhou (2003), Charlin, Poupart, and Shioda (2007), Foka and Trahanias (2007), Theocharous and Kaelbling (2003) and Kurniawati et al. (2009). Pineau et al.'s PolCA+ (2003b) algorithm uses a hierarchical approach to solving discrete-state POMDPs. Similarly, Hansen and Zhou (2003) propose hierarchical controllers that exploit a user-specified hierarchy for planning, while Charlin et al. (2007) provide a method for automatically discovering a problem hierarchy. Yu, Chuang, Gerkey, Gordon and Ng (2005) provide an optimal algorithm for planning if no observations were available. Foka and Trahanias's (2007) solution involves building a hierarchy of nested representations and solutions. Their focus is on discrete-state problems, particularly navigation applications. Theocharous and Kaelbling's (2003) discrete-state reinforcement learning approach samples observation trajectories and solves for the expected reward of a discrete set of belief points using function approximation. Kurniawati et al. (2009) recently used macro-actions to guide the sampling of belief points for use in an offline point-based POMDP solver.

However, these prior macro-action POMDP approaches compute a value function off-line, are not aimed at scaling to very large domains, and will struggle in the environments considered in this paper. An exception to this is the work by Hsiao and colleagues (2008, 2010) who used a form of macro-actions for those robot manipulation tasks that involve a large state space. The focus of their work is on robust manipulation under uncertainty, and their work only considers a very short horizon of action trajectories. Except for the work by Kurniawati et al. (2009), all these macro-action POMDP approaches, like our PBD algorithm, assume the macro-actions are provided by a domain expert.

In the sensor resource management domain, planning under uncertainty techniques are used in the context of planning sensor placements to track single or multiple targets. Existing algorithms often adopt a myopic, or greedy strategy when it comes to planning (Krause & Guestrin, 2007), but notable exceptions include the work by Scott et al. (2009) and Kreucher, Hero III, Kastella, and Chang (2004). Kreucher et al. describe a multi-target tracking problem, where non-myopic sensor management is necessary for multi-target tracking. The authors use a particle filter approach to represent the agent's belief of the target's location, and seek to find paths that will result in the greatest KL divergence in density before and after the measurement. To look ahead more than one action, this algorithm uses Monte Carlo sampling to generate possible observation outcomes. They also provide an information-directed path searching scheme to reduce the complexity of the Monte Carlo sampling, as well as value heuristics that will help direct the search. It is possible that some of their insights could be used in combination with our macro-action formulation to strengthen both approaches. In the experimental section we compared our approach to the work by Scott et al. (2009), who directly formulated target tracking as a POMDP, and proposed the Nominal Belief Optimization (NBO) algorithm that computes the most likely belief after an action for deeper forward search. In contrast, our algorithm explicitly computes the entire set of possible posterior beliefs after a macro-action. Recently two groups (Erez & Smart, 2010; Platt, Tedrake, Lozano-Perez, & Kaelbling, 2010) have independently proposed an approach that lies in the middle of this spectrum: beliefs are updated by assuming that the most likely observation is received, but the variance is increased. In contrast, our approach represents that each resulting belief may be fairly peaked, but the mean of the beliefs may be spread out. This more complete representation may be advantageous if there are sharp changes in the reward function.

As stated in the introduction, the finite-horizon forward search, act, and re-plan strategy PBD follows can be seen as an instance of the Model Predictive Control/Receding Horizon Control (MPC/ RHC) framework from the controls community. Examples of MPC and RHC include the work by Kuwata and How (2004), Bellingham, Richards, and How (2002), and Richards, Kuwata, and How (2003). A special case of RHC control is Certainty Equivalence Control, or CEC (see Bertsekas, 2007 for an overview). In fully observable systems, CEC first assumes all stochastic operations (such as transitions) take on their expected value, and then solves a finite-horizon deterministic control problem. CEC may be applied in partially observable environments by first sampling an initial state from the belief state. Though CEC can be very efficient in large domains, a key limitation of its use in partially observable environments is that a CEC-style controller will never take information-gathering actions. Returning to the generic class of MPC approaches, to our knowledge no prior model predictive controllers have used macro-actions nor developed the notion of a pos-terior distribution of beliefs, which enables our PBD approach to scale to large uncertain domains where a multi-step lookahead is required.

## 8. Conclusion

In this paper we have presented the Posterior Belief Distribution algorithm. PBD is a forward-search algorithm for large (consisting of many variables, each of which can take on many values) partially observable domains. PBD analytically and efficiently computes the resulting distribution of posterior belief states possible after a sequence of actions. This allows the computational cost of evaluating the reward associated with a macro-action to be tractable, which we leverage to enable longer horizon lookahead search during online planning. We have presented theoretical and experimental results evaluating the performance and computational cost of our macro-action algorithms. Our algorithms were applied to problem domains that span multiple research communities, and consistently performed better than prior approaches in large domains which require multi-step lookahead for good performance. Finally, we demonstrated our algorithm on a real robotic helicopter, underscoring the applicability of our algorithm for planning in real-world, long-horizon, partially observable domains.

## 9. Acknowledgments

Ruijie He, E. Brunskill and N. Roy were supported by the National Science Foundation (NSF) Division of Information and Intelligent Systems (IIS) under Grant #0546467 and by the Office of Naval Research under the “Decentralized Reasoning in Reduced Information Spaces” project, Contract # N00014-09-1-1052.

We wish to thank Finale Doshi-Velez, Alborz Geramifard, Josh Joseph, Brandon Luders, Javier Velez, and Matthew Walter for valuable discussions and feedback. Daniel Gurdan, Jan Stumpf and Markus Achtelik provided the quadrotor helicopter and the support of Ascending Technologies. Abraham Bachrach, Anton De Winter, Garrett Hemann, Albert Huang, and Samuel Prentice assisted with the development of the software and hardware for the helicopter demonstration. We also appreciate early POMDP forward search discussions with Leslie Pack Kaelbling and Tomas Lozano-Perez.

## Appendix A: Exponential Family Kalman Filter

Building on statistical economics research for time-series analysis of non-Gaussian observations (Durbin & Koopman, 2000), we present the Kalman filter equivalent for systems with linear-Gaussian state-transitions and observation models that belong to the exponential family of distributions.

The state-transition and observation models can be represented as follows:

$$s_t = A_t s_{t-1} + B_t a_t + \varepsilon_t, \qquad s_{t-1} \sim N(\mu_{t-1}, \Sigma_{t-1}), \qquad \varepsilon_t \sim N(0, P_t) \tag{68}$$

$$p(z_t|\theta_t) = \exp(z_t^T \theta_t - \beta_t(\theta_t) + \kappa_t(z_t)), \qquad \theta_t = W(s_t). \tag{69}$$

For the state-transition model, $s_t$ is the system's hidden state, $a_t$ is the control actions, $A_t$ and $B_t$ are the linear transition matrices, and $\epsilon_t$ is the state-transition Gaussian noise with covariance $P_t$.

The observation model belongs to the exponential family of distributions. $\theta_t$ and $\beta_t(\theta_t)$ are the canonical parameter and normalization factor of the distribution, and $W(.)$ maps the states to canonical parameter values. $W(.)$ depends on the particular member of the exponential family. For

ease of notation, we let

$$\upsilon_t(z_t|\theta_t) = -\log p(z_t|\theta_t) = -z_t^T\theta_t + \beta_t(\theta_t) + \kappa_t(z_t). \tag{70}$$

Following the traditional Kalman filter, the process update can be written as

$$\overline{\mu}_t = A_t\mu_{t-1} + B_t a_t, \qquad \overline{\Sigma}_t = A_t\Sigma_{t-1}A_t^T + P_t, \tag{71}$$

where $\overline{\mu}_t$ and $\overline{\Sigma}_t$ are the mean and covariances of the posterior belief after the process update but before the measurement udpate. For the measurement update, we seek to find the conditional mode

$$\begin{aligned}
\mu_t &= \arg\max_{s_t} p(s_t|z_t) && (72)\\
&= \arg\max_{s_t} p(z_t|s_t)\overline{b}(s_t) \qquad \text{(Bayes rule)} && (73)\\
&= \arg\max_{s_t} p(z_t|\theta_t)\overline{b}(s_t) && (74)\\
&= \arg\max_{s_t} \exp(-J_t),\ \text{where } J_t = -\log p(z_t|\theta_t) + \frac{1}{2}(s_t-\overline{\mu}_t)^T\overline{\Sigma}_t^{-1}(s_t-\overline{\mu}_t) && (75)\\
\Rightarrow \qquad 0 &= \frac{\partial J_t}{\partial s_t}\Big|_{s_t=\mu_t} = \frac{\partial \upsilon_t(z_t,\theta_t)}{\partial\theta_t}\frac{\partial\theta_t}{\partial s_t} + \overline{\Sigma}_t^{-1}(\mu_t-\overline{\mu}_t). && (76)
\end{aligned}$$

Taking the derivative of $\theta_t = W(s_t)$ about the prior mean $\overline{\mu}_t$, we let

$$Y_t = \frac{\partial W(s_t)}{\partial s_t}\Big|_{s_t=\overline{\mu}_t}. \tag{77}$$

Similarly, performing a Taylor expansion on $\frac{\partial\upsilon_t(z_t|\theta_t)}{\partial\theta_t}$ about $\overline{\theta}_t = W(\overline{\mu}_t)$,

$$\begin{aligned}
\frac{\partial\upsilon_t(z_t|\theta_t)}{\partial\theta_t} &= \frac{\partial\upsilon_t(z_t|\theta_t)}{\partial\theta_t}\Big|_{\theta_t=\overline{\theta}_t} + \frac{\partial^2\upsilon_t(z_t|\theta_t)}{\partial\theta_t\partial\theta_t^T}\Big|_{\theta_t=\overline{\theta}_t}(\theta_t-\overline{\theta}_t) && (78)\\
\frac{\partial\upsilon_t(z_t|\theta_t)}{\partial\theta_t} &= \dot{\upsilon}_t + \ddot{\upsilon}_t(\theta_t-\overline{\theta}_t) && (79)\\
\text{where} \qquad \dot{\upsilon}_t &= \frac{\partial}{\partial\theta_t}(-z_t^T\theta_t + \beta_t(\theta_t) - \kappa_t(z_t))\Big|_{\theta_t=\overline{\theta}_t}, \qquad \text{(Eqn. 70)} && (80)\\
&= \frac{\partial\beta_t(\theta_t)}{\partial\theta_t}\Big|_{\theta_t=\overline{\theta}_t} - z_t && (81)\\
\dot{\upsilon}_t &= \dot{\beta}_t - z_t && (82)\\
\text{and} \qquad \ddot{\upsilon}_t &= \frac{\partial^2\beta_t(z_t|\theta_t)}{\partial\theta_t\partial\theta_t^T}\Big|_{\theta_t=\overline{\theta}_t}(\theta_t-\overline{\theta}_t). && (83)\\
\ddot{\upsilon}_t &= \ddot{\beta}_t && (84)
\end{aligned}$$

Plugging Equations 82 and 84 into Equation 79, and then into Equation 76,

$$\begin{aligned}
Y_t^T\ (\dot{\beta}_t - z_t + \ddot{\beta}_t(\theta_t-\overline{\theta}_t)) &= -\ \overline{\Sigma}_t^{-1}(\mu_t-\overline{\mu}_t) && (85)\\
Y_t^T\ \ddot{\beta}_t(\ddot{\beta}_t^{-1}(\dot{\beta}_t - z_t) - \overline{\theta}_t + \theta_t) &= -\ \overline{\Sigma}_t^{-1}(\mu_t-\overline{\mu}_t) && (86)\\
Y_t^T\ \ddot{\beta}_t((\overline{\theta}_t - \ddot{\beta}_t^{-1}(\dot{\beta}_t - z_t)) - \theta_t) &= \overline{\Sigma}_t^{-1}(\mu_t-\overline{\mu}_t) && (87)\\
Y_t^T\ \ddot{\beta}_t(\tilde{z}_t - W(s_t)) &= \overline{\Sigma}_t^{-1}(\mu_t-\overline{\mu}_t), && (88)
\end{aligned}$$

where $\tilde{z}_t = (\overline{\theta}_t - \ddot{\beta}_t^{-1}(\dot{\beta}_t - z_t))$ is the projection of the observation onto the parameter space of the exponential family distribution, and is independent of $s_t$. In Equation 88 we substituted $\theta_t$ using Equation 69.

### Mean Update

Using Equation 88 and substituting $\mu_t$ for $s_t$,

$$\overline{\Sigma}_t^{-1}(\mu_t - \overline{\mu}_t) = Y_t^T\ \ddot{\beta}_t(\tilde{z}_t - W(\mu_t)) \tag{89}$$

$$= Y_t^T\ \ddot{\beta}_t(\tilde{z}_t - W(\mu_t)) + W(\overline{\mu}_t) - W(\overline{\mu}_t) \tag{90}$$

$$= Y_t^T\ \ddot{\beta}_t(\tilde{z}_t - W(\overline{\mu}_t)) - Y_t^T\ \ddot{\beta}_t(W(\mu_t) - W(\overline{\mu}_t)). \tag{91}$$

Linearizing $W(s_t)$ about $\overline{\mu}_t$,

$$W(s_t) = W(\overline{\mu}_t) + W'(s_t)_{s_t=\overline{\mu}_t}(s_t - \overline{\mu}_t) \tag{92}$$

$$= W(\overline{\mu}_t) + Y_t\ (\mu_t - \overline{\mu}_t) \tag{93}$$

$$\Rightarrow \qquad \overline{\Sigma}_t^{-1}(\mu_t - \overline{\mu}_t) = Y_t^T\ \ddot{\beta}_t(\tilde{z}_t - W(\overline{\mu}_t)) - Y_t^T\ \ddot{\beta}_t Y_t\ (\mu_t - \overline{\mu}_t) \tag{94}$$

$$Y_t^T\ \ddot{\beta}_t(\tilde{z}_t - W(\overline{\mu}_t)) = (\overline{\Sigma}_t^{-1} + Y_t^T\ \ddot{\beta}_t Y_t\ )(\mu_t - \overline{\mu}_t) \tag{95}$$

$$= \Sigma_t^{-1}(\mu_t - \overline{\mu}_t) \tag{96}$$

$$\Rightarrow \qquad \mu_t - \overline{\mu}_t = \Sigma_t Y_t^T\ \ddot{\beta}_t(\tilde{z}_t - W(\overline{\mu}_t)), \tag{97}$$

where $\Sigma_t Y_t^T\ \ddot{\beta}_t = \tilde{K}_t$ is the Kalman gain for non-Gaussian exponential family distributions. Via a standard transformation, the Kalman gain can be written in terms of covariances other than $\Sigma_t$,

$$\tilde{K}_t = \overline{\Sigma}_t Y_t^T\ (Y_t \overline{\Sigma}_t Y_t^T + \ddot{\beta}_t^{-1})^{-1} \tag{98}$$

$$\text{and} \qquad \mu_t = \overline{\mu}_t + \tilde{K}_t(\tilde{z}_t - W(\overline{\mu}_t)). \tag{99}$$

### Covariance Update

Given a Gaussian posterior belief, $\frac{\partial^2 J}{\partial s_t^2}$ is the inverse of the covariance of the agent's belief

$$\Sigma_t^{-1} = \frac{\partial^2 J}{\partial s_t^2} \tag{100}$$

$$= \frac{\partial}{\partial x}(\overline{\Sigma}_t^{-1}(s_t - \overline{\mu}_t) - Y_t^T\ \ddot{\beta}_t(\tilde{z}_t - W(s_t))) \tag{101}$$

$$= \overline{\Sigma}_t^{-1} + Y_t^T\ \ddot{\beta}_t Y_t \tag{102}$$

$$\Rightarrow \Sigma_t = (\overline{\Sigma}_t^{-1} + Y_t^T\ \ddot{\beta}_t Y_t)^{-1}. \tag{103}$$

## Appendix B. Rock Sample Observation Model

In the Rocksample problem, the Bernoulli observation function can be written as follows. Recall that $r_t$ is the agent's position at time $t$, $RB_i$ is the location of the information beacon associated with rock $i$, $z_{i,t}$ is a binary observation of the value of rock $i$ at time $t$, and $s_{i,t}$ is the true value of rock $i$ at time $t$. Then if we let $d_{i,t} = \| r_t - RB_i \|_2$, then

$$\begin{aligned} & p(z_{i,t}|RV_{i,t} = s_{i,t}, r_t, RB_i) && (104)\\ &\quad = (0.5 + (s_{i,t} - 0.5)2^{-d_{i,t}/D_0})^{z_{i,t}}(0.5 - (s_{i,t} - 0.5)2^{-d_{i,t}/D_0})^{1-z_{i,t}} && (105)\\ &\quad = \exp(z_{i,t} \ln \frac{0.5 + (s_{i,t} - 0.5)2^{-d_{i,t}/D_0}}{0.5 - (s_{i,t} - 0.5)2^{-d_{i,t}/D_0}} + \ln(0.5 - (s_{i,t} - 0.5)2^{-d_{i,t}/D_0})) && (106)\\ &\quad = \exp(z_{i,t}\theta_t - \beta_t(\theta_t)). && (107) \end{aligned}$$

We therefore have the parameters of the exponential family observation model

$$\begin{aligned} \theta_{i,t} &= W(s_{i,t}, r_t, RB_i) && (108)\\ &= \ln \frac{0.5 + (s_{i,t} - 0.5)2^{-d_{i,t}/D_0}}{0.5 - (s_{i,t} - 0.5)2^{-d_{i,t}/D_0}} && (109)\\ \beta_{i,t} &= -\ln(0.5 - (s_{i,t} - 0.5)2^{-d_{i,t}/D_0}) && (110)\\ &= \ln(\exp(\theta_{i,t}) + 1). && (111) \end{aligned}$$

We can then derive the derivatives $Y_{i,t}$ and $\ddot{\beta}_{i,t}$

$$\begin{aligned} Y_t &= \left.\frac{\partial W(s_{i,t}, r_t, RB_i)}{\partial s_{i,t}}\right|_{s_{i,t}=\hat{m}_{i,t}} && (112)\\ &= \frac{\partial}{\partial s_{i,t}} \ln \left.\frac{0.5 + (s_{i,t} - 0.5)2^{-d_{i,t}/D_0}}{0.5 - (s_{i,t} - 0.5)2^{-d_{i,t}/D_0}}\right|_{s_{i,t}=\hat{m}_{i,t}} && (113)\\ &= \frac{2^{-d_{i,t}/D_0}}{0.5 + (\hat{m}_{i,t} - 0.5)2^{-d_{i,t}/D_0}} \cdot \frac{1}{0.5 - (\hat{m}_{i,t} - 0.5)2^{-d_{i,t}/D_0}} && (114) \end{aligned}$$

where $\hat{s}_{i,t}$ is the mean of the belief used for linearization. Since

$$\Rightarrow \qquad \beta_{i,t} = \ln(\exp(\theta_{i,t}) + 1), \qquad (115)$$

then

$$\begin{aligned} \ddot{\beta}_{i,t} &= \left.\frac{\partial^2 b_{i,t}}{\partial \theta_{i,t}^2}\right|_{\theta_{i,t}=\hat{\theta}_{i,t}} && (116)\\ &= \frac{\exp(\hat{\theta}_{i,t})}{\exp(\hat{\theta}_{i,t}) + 1} - \frac{\exp(2\hat{\theta}_{i,t})}{(\exp(\hat{\theta}_{i,t}) + 1)^2}. && (117) \end{aligned}$$

## Appendix C. Target Tracking Observation Model

We adopt an observation model for target tracking where the target observation obtained has Gaussian noise and the noise covariance $\Sigma_{zi}$ is a function of the position of the helicopter and target $i$:

$$\begin{aligned} \begin{bmatrix} z_{xi} \\ z_{yi} \\ z_{\theta i} \end{bmatrix} &= f\left(\begin{bmatrix} x_i \\ y_i \\ \theta_i \end{bmatrix}\right) + \mathcal{N}(0, \Sigma_{zi}) \\ \Sigma_{zi} &= g(x_i, y_i, x_a, y_a, h_a), \end{aligned}$$

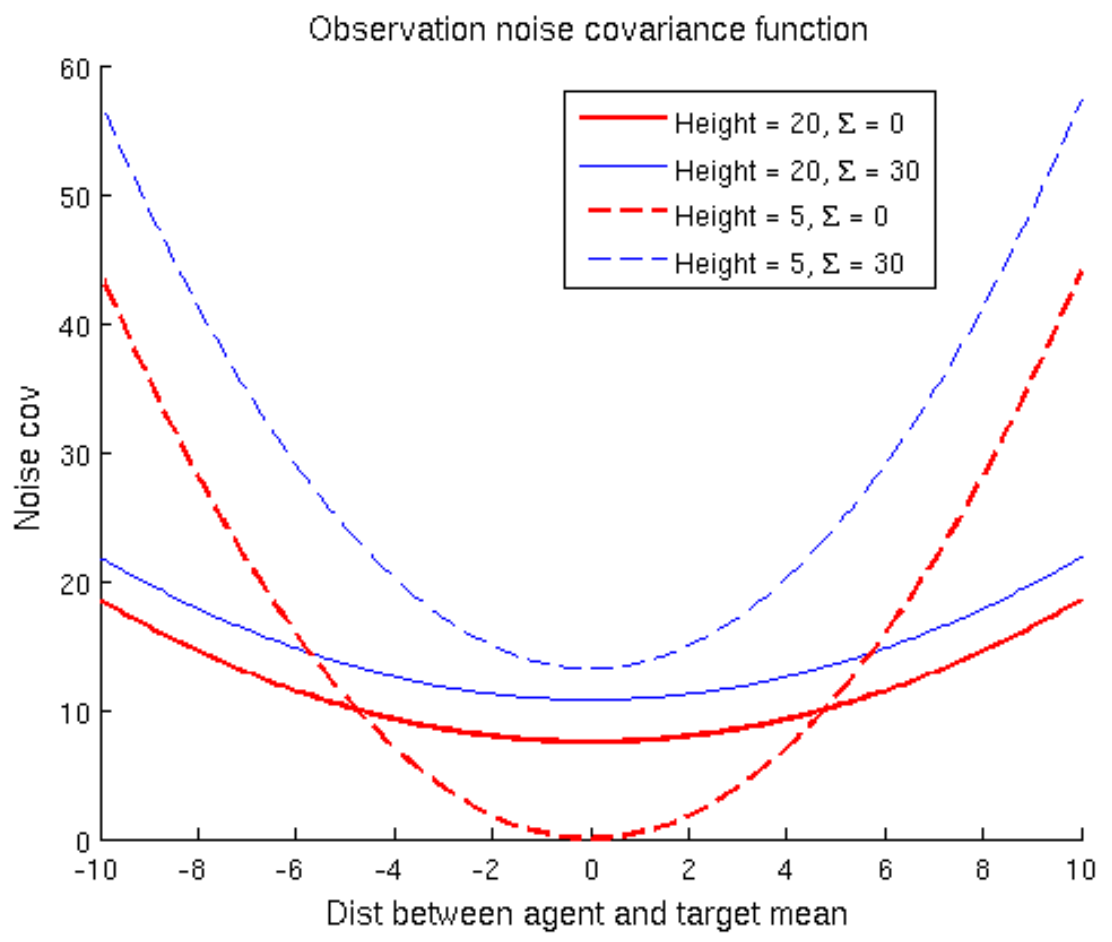

Figure 11: The observation noise covariance is a function of the height of helicopter, the distance between the helicopter and the mean of the target belief, and the covariance of the target belief. At lower altitudes, the helicopter can make better observations of targets close to it, but has a limited field of vision. At higher heights, the helicopter can see a larger area but even close targets are more noisily observed.

where $x_i, y_i, \theta_i$ is the pose of target $i$, while $x_a, y_a, h_a$ correspond to the agent's position and height in the environment. $z_{xi}, z_{yi}, z_{\theta_i}$ is the observation of target $i$ in image coordinates.

The covariance function itself is specified as

$$g(x_i, y_i, x_a, y_a, h_a) = C_1 h_a + C_2 \frac{\left(\begin{bmatrix} x_i \\ y_i \end{bmatrix} - \begin{bmatrix} x_a \\ y_a \end{bmatrix}\right)\left(\begin{bmatrix} x_i \\ y_i \end{bmatrix} - \begin{bmatrix} x_a \\ y_a \end{bmatrix}\right)^T}{h_a} + C_3,$$

where $C_1$, $C_2$ and $C_3$ are constants.

In the generic belief update expression where the target position, $s_i = [x_i; y_i; \theta_i]$, is unknown,

$$b'(s'_i) \propto p(z|s'_i, a, \Sigma_{zi}) \int_{s_i} p(s'_i|s_i, a) b(s_i) ds_i \quad s.t. \quad \int_{s'_i} b'(s'_i) ds'_i = 1,$$

which means that each possible $s'_i$ would be associated with a different covariance $\Sigma_{zi}$. Performing this integration exactly would not keep the distribution Gaussian. Instead, we approximate the observation model by computing a single expected covariance $\hat{\Sigma}_{zi}$ given the current belief distribution:

$$\hat{\Sigma}_{zi} = E[\Sigma_{zi}] = \int_{s_i} b(s_i) \Sigma_{zi}(s_i) ds_i.$$

Substituting in the exact expressions for the covariance function and the belief after an action is taken but before incorporating the measurement, $b^a(s) \sim \mathcal{N}(s_i|\overline{\mu}, \overline{\Sigma})$, we get:

$$E[\Sigma_{zi}] = \int \mathcal{N}\left(\begin{bmatrix} x_i \\ y_i \end{bmatrix} \middle| \overline{\mu}_{xy}, \overline{\Sigma}_{xy}\right)\left(C_1 h_a - \frac{C_2}{h_a}\left(\begin{bmatrix} x_i \\ y_i \end{bmatrix} - \begin{bmatrix} x_a \\ y_a \end{bmatrix}\right)\left(\begin{bmatrix} x_i \\ y_i \end{bmatrix} - \begin{bmatrix} x_a \\ y_a \end{bmatrix}\right)^T + C_3\right) dx_i dy_i.$$

and by adding and subtracting $\overline{\mu}_{xy}$ from the second term, reduces to

$$E[\Sigma_{zi}] \ = C_1 h_a + \frac{C_2}{h_a}\left(\overline{\mu}_{xy} - \begin{bmatrix} x_a \\ y_a \end{bmatrix}\right)\left(\overline{\mu}_{xy} - \begin{bmatrix} x_a \\ y_a \end{bmatrix}\right)^T + \frac{C_2}{h_a}\overline{\Sigma}_{xy}$$

where $\overline{\mu}_{xy}, \overline{\Sigma}_{xy}$ refer to the translational components of the agent's belief.

In contrast to simpler observation models, our observation model has the desirable characteristic that if a target's location is very uncertain, namely its covariance $\overline{\Sigma}_{xy}$ is very large, then even if the target's mean location is close to the helicopter's mean location, the expected benefit of receiving an observation (in terms of reducing the target's uncertainty) is still small. This property comes out automatically from the above derivation, since $E[\Sigma_{zi}]$ includes the current target covariance $\overline{\Sigma}_{xy}$. Figure 11 provides an illustration of the expected covariance for different locations of the target relative to the agent, agent heights, and target belief covariances.